\documentclass{sigkddExp}
\usepackage{booktabs} 
\usepackage{enumerate}
\usepackage{amsmath}
\usepackage{amsfonts}
\usepackage{graphicx}
\usepackage[ruled,vlined]{algorithm2e}
\usepackage{algpseudocode}
\usepackage{bm}
\usepackage{silence}
\usepackage{subfigure}
\usepackage{multirow}
\usepackage{array}
\usepackage{xcolor}
\usepackage{amssymb}
\usepackage{enumitem}
\usepackage{makecell}

\begin{document}
%

\title{Data Augmentation for Deep Graph Learning: \\A Survey}

\author{
Kaize Ding$^{1}$,~~~Zhe Xu$^{2}$,~~~Hanghang Tong$^{2}$,~~~Huan Liu$^{1}$\\
\affaddr{$^1$Arizona State University}\\
\affaddr{$^2$University of Illinois Urbana-Champaign}\\

\normalsize{kaize.ding@asu.edu, zhexu3@illinois.edu,  htong@illinois.edu, huanliu@asu.edu}

}


\maketitle

\begin{abstract}
Graph neural networks, a powerful deep learning tool to model graph-structured data, have demonstrated remarkable performance on numerous graph learning tasks. To address the data noise and data scarcity issues in deep graph learning, the research on graph data augmentation has intensified lately. However, conventional data augmentation methods can hardly handle graph-structured data which is defined in non-Euclidean space with multi-modality. In this survey, we formally formulate the problem of graph data augmentation and further review the representative techniques and their applications in different deep graph learning problems. Specifically, we first propose a taxonomy for graph data augmentation techniques and then provide a structured review by categorizing the related work based on the augmented information modalities. Moreover, we summarize the applications of graph data augmentation in two representative problems in data-centric deep graph learning: (1) reliable graph learning which focuses on enhancing the utility of input graph as well as the model capacity via graph data augmentation; and (2) low-resource graph learning which targets on enlarging the labeled training data scale through graph data augmentation. For each problem, we also provide a hierarchical problem taxonomy and review the existing literature related to graph data augmentation. Finally, we point out promising research directions and the challenges in future research.
\end{abstract}


\section{Introduction}




Graphs have been widely used for modeling a plethora of structured or relational systems, such as social networks, knowledge graphs, and academic graphs, where nodes represent the entities and edges denote the relations between entities. As a powerful deep learning tool to distill the knowledge behind graph-structured data, graph neural networks (GNNs) which generally follow a recursive message-passing scheme, have drawn a surge of research interest lately. Owing to its state-of-the-art performance, deep graph learning (DGL) nowadays has achieved remarkable success in a wide spectrum of graph analytical tasks~\cite{ding2019deep,wang2020next,ding2020more}.



Despite the superb power of GNNs, their effectiveness in DGL largely depends on high-quality input training graph(s) and ground-truth labels. The performance of GNNs tends to be delicate on real-world graphs, mainly because of their incapability of handling the following challenges: (1) On the one hand, prevailing DGL models are predominantly designed for the supervised or semi-supervised setting where sufficient ground-truth labels are available~\cite{ding2020graph,ding2022meta}. Considering the fact that data labeling on graphs is always time-consuming, labor-intensive, and rarely complete, the overreliance on labeled data poses great challenges to DGL models in real-world scenarios. Meanwhile, there are increasingly more tasks and domain-specific applications that are low-resource, having a paucity of labeled training examples. When ground-truth labels are extremely scarce, DGL models may easily overfit and be hard to generalize, losing their efficacy in solving various downstream DGL tasks~\cite{sun2020multi,ding2022meta}. (2) On the other hand, real-world graphs are usually extracted from complex interaction systems which inevitably contain redundant, erroneous, or missing features and connections. In addition, the noxious manipulations from adversaries as well as the inherent limitations of GNNs such as the oversmoothing issue~\cite{li2018deeper} also bring additional challenges to the success of reliable DGL. Directly training GNN-based models on such inferior graphs that are not clean and consistent with the properties of GNNs might lead to serious performance degradation~\cite{dai2021nrgnn}.




To improve the sufficiency and quality of training data, data augmentation is proposed as an effective tool to augment the given input data by either slightly modifying existing data instances or generating synthetic instances from existing ones. The importance of data augmentation has been well recognized in the computer vision~\cite{shorten2019survey} and natural language processing domains~\cite{zhao2021data} in the past few years. More recently, data augmentation techniques have also been explored in the graph domain to push forward the performance boundary of DGL and demonstrated promising results~\cite{zhao2021data,qiu2020gcc}. Apart from conventional image or text data, graph-structured data is known to be far more complicated with heterogeneous information modalities and complex graph properties, yielding a broader design space as well as the additional challenges for graph data augmentation (GraphDA). Though this line of research has been actively conducted lately, the problem of GraphDA has not been well formulated and researchers commonly adopt GraphDA techniques (e.g., edge perturbation, feature masking) arbitrarily without clear preference. Hence, it poses great challenge for researchers to grasp the design principles behind and further leverage them to solve specific DGL problems. Therefore, a timely and systematic review of GraphDA is of great benefit to be aware of existing research in this field and what the challenges are for conducting future research.


\smallskip
\noindent\textbf{Contributions.}  In this work, we present a forward-looking and up-to-date survey for GraphDA and its applications in solving data-centric DGL problems. In summary, our major contributions are as follows:
\begin{itemize}[leftmargin=*,noitemsep,topsep=1.5pt]
    \item To the best of our knowledge, this is the first survey for GraphDA. We provide a formal formulation for this emerging research area and review the related recent advances, which can facilitate the understanding of important issues to promote future research.
    
    \item We present a comprehensive taxonomy of GraphDA which categorizes the existing techniques in terms of the target augmentation modality (i.e., feature-oriented, structure-oriented, and label-oriented) and provides a clear design space for developing and customizing new GraphDA methods for different DGL problems. 
    
    \item We discuss the applications of GraphDA in solving two major research problems in data-centric DGL, i.e., optimal graph learning and low-resource graph learning, and review the prevalent learning paradigms for solving specific sub-problems. We also outline the open issues and promising future directions in this area.
\end{itemize}

\smallskip
\noindent\textbf{Connection to Existing Surveys.} Although there are a few survey papers~\cite{liu2022graph,wu2021self,xie2022self} have related content to GraphDA, those surveys predominately focus on a single DGL problem such as graph self-supervised learning or graph adversarial defense. Other data-centric DGL problems related to GraphDA are largely overlooked and the corresponding GraphDA techniques are rarely discussed. In contrast, our survey includes a detailed and systematic review of GraphDA techniques and their corresponding applications for solving two most representative data-centric DGL problems. Meanwhile, we also discuss a list of under-explored directions in GraphDA, which can shed great light on future DGL research.

\smallskip
\noindent\textbf{Structure Overview.} This survey is structured as follows. Section \ref{sec:pre} first gives background on graph neural networks (GNNs) and deep graph learning. Then in Section \ref{sec:taxonomy} we provide a comprehensive taxonomy of GraphDA techniques based on the focused augmentation modality of the input graph(s). In the following two sections, we describe the applications of GraphDA techniques for solving two data-centric DGL problems, i.e., \textit{Low-resource Graph Learning} (Section \ref{sec: low-resource application}) and \textit{Reliable Graph Learning} (Section \ref{sec: optimal graph learning}). Specifically, Section \ref{sec: low-resource application} includes the techniques based on GraphDA for solving graph self-supervised learning and graph semi-supervised learning. Section \ref{sec: optimal graph learning} covers the content of improving the robustness, expressiveness, and scalability of DGL models from the data augmentation perspective. Within each subsection, we introduce the methods grouped by their related GraphDA techniques. Finally, Section \ref{sec:future challenges} discusses challenges and future directions in GraphDA.

\section{Preliminaries}
\label{sec:pre}

\subsection{Notations and Definitions}
We briefly introduce the main symbols and notations used throughout this paper. We use bold uppercase letters for matrices (e.g., $\mathbf{X}$), bold lowercase letters for vectors (e.g., $\mathbf{v}$), lowercase and uppercase letters for scalars (e.g., $d$, $n$), and calligraphic letters for sets (e.g., $\mathcal{N}$). We use $\mathbf{X}[i,j]$ to represent the entry of matrix $\mathbf{X}$ at the $i$-th row and the $j$-th column, $\mathbf{X}[i,:]$ to represent the $i$-th row of matrix $\mathbf{X}$, and $\mathbf{X}[:,j]$ to represent the $j$-th column of matrix $\mathbf{X}$. Similarly, $\mathbf{v}[i]$ denotes the $i$-th entry of vector $\mathbf{v}$.



For the general purpose, we focus on undirected attributed graphs in this survey. A graph can be represented as $\mathcal{G}=\{\mathcal{V},\mathcal{E}\}$ where $\mathcal{V}$ denotes the node set $\{v_i\}_{i=1}^n$ and $\mathcal{E}$ denotes the edge set $\{e_i\}_{i=1}^m$. In matrix form, it can also be represented as $\mathcal{G}=(\mathbf{A},\mathbf{X})$, where $\mathbf{A}\in\mathbb{R}^{n\times n}$ denotes the adjacency matrix and $\mathbf{X}\in\mathbb{R}^{n\times d}$ denotes the node feature matrix. Here $n$ is the number of nodes, $m$ is the number of edges, and $d$ is the feature dimension. For supervised tasks, a part of the node labels $\mathbf{y}\in\mathbb{R}^{n}$, the edge labels $\mathbf{Y}\in\mathbb{R}^{n\times n}$, and the graph label $y$ are provided. For tasks with multiple graphs (e.g., graph-level tasks), we appropriately use subscripts to describe graphs and corresponding components. For example, $\mathbf{A}_{i}$ denotes the adjacency matrix of the $i$-th graph $\mathcal{G}_{i}$.


\subsection{Graph Neural Networks}
Graph neural networks (GNNs)~\cite{wu2020comprehensive}\cite{zhang2019graph} are the extension of the neural network models~\cite{bishop1994neural} onto graph data. They show great flexibility and strong expressiveness to extract representations of various graph components and are becoming the core modules of broad graph learning tasks. In this subsection, we will briefly introduce the general message passing formulas~\cite{gilmer2017neural} of the mainstream GNNs as they are widely adopted by the works remaining of this paper.

The message passing framework of GNNs can be mathematically presented as follows:
\begin{subequations}
\begin{align}
    \mathbf{m}_i^{t+1} &= \sum_{j\in N(i)}\texttt{Message}(\mathbf{h}_i^t, \mathbf{h}_j^t),\label{eq:mpnn message}\\
    \mathbf{h}_i^{t+1} &= \texttt{Update}(\mathbf{h}_i^t, \mathbf{m}_i^{t+1}),\label{eq:mpnn update}
\end{align}
\end{subequations}
\noindent where $\mathbf{h}_i^t$ denotes the representation of the node $i$ at $t$-th layer and $\mathbf{m}_i^{t+1}$ denotes the message aggregates on the node $i$ at the $(t+1)$-th layer. The initial node representation is the node feature (i.e., $\mathbf{h}_i^0=\mathbf{X}[i]$) and the node neighborhood can be represented by the adjacency matrix (i.e., $\mathbf{A}[i,j]=1$ is equivalent to $j\in N(i)$). For graph-level tasks, graph representation $\mathbf{h}_{G}$ can be obtained through a $\texttt{Readout}$ function as follows:
\begin{equation}
    \mathbf{h}_{G} = \texttt{Readout}(
    \mathbf{H}),
\end{equation}
\noindent where $\mathbf{H}$ can be the node representations from the final layer or any intermediate layer.


\subsection{Deep Graph Learning Tasks}


In this subsection, we introduce several mainstream DGL tasks on which GraphDA techniques are widely used. We categorize tasks according to their objective graph components (i.e., node, edge, and graph).

\smallskip
\noindent\textbf{Node-level DGL Tasks.} 
Node-level DGL tasks aim to find a mapping $p_{\phi}$ from the given graphs to node properties by minimizing a utility loss $L_\texttt{util}$ as follows:
\begin{equation*}
    \phi^* = \arg \min_{\phi} L_\texttt{util}\Big(p_{\phi}(\mathcal{G}), \mathbf{y}\Big),
\end{equation*}
whose typical example is \emph{semi-supervised node classification}. Given the labels of partial nodes for training, the goal is to predict the labels of (a part of) unlabelled nodes. A classic implementation of a node classifier is a node encoder (e.g., GNNs) working with a multi-class classifier (e.g., an MLP).

\smallskip
\noindent\textbf{Edge-level DGL Tasks.} Edge-level DGL tasks focus on finding a mapping $p_{\phi}$ from the given graphs to edge properties which can be presented as below:
\begin{equation*}
    \phi^* = \arg \min_{\phi} L_\texttt{util}\Big(p_{\phi}(\mathcal{G}), \mathbf{Y}\Big).
\end{equation*}
Taking \emph{link prediction} as an example, its goal is to discriminate if there is an edge between specified node pair. A common implementation is a binary GNN classifier whose input is the edge embeddings (e.g., the aggregation of node embeddings of the head and tail nodes).

\smallskip
\noindent\textbf{Graph-level DGL Tasks.} A graph-level task considers every graph as a data sample and infers the property of graph(s) by a mapping $p_{\phi}$. Its mathematical formulation is as follows:
\begin{equation*}
    \phi^* = \arg \min_{\phi} L_\texttt{util}\Big(\{p_{\phi}(\mathcal{G}_i)\}, \{y_i\}\Big).
\end{equation*}
For instance,
in the \emph{graph classification} task, some labeled graphs are provided and the goal is to predict the labels of graphs of interest. A general solution is to aggregate the node embeddings into a graph embedding via a readout function and feed the graph embeddings into a classifier.

\section{Techniques of Graph Data Augmentation}
\label{sec:taxonomy}
The goal of graph data augmentation (GraphDA) is to find a transformation function $f_\theta (\cdot): \mathcal{G}\rightarrow \tilde{\mathcal{G}}$ to generate augmented graph(s) $\{\tilde{\mathcal{G}}_i=(\tilde{\mathbf{A}}_i,\tilde{\mathbf{X}}_i)\}$ that can enrich or enhance the preserved information from the given graph(s). In terms of whether the parameter $\theta$ can be updated or not during the learning process, most, if not all, of the GraphDA methods can be classified to: \textit{non-learnable} and \textit{learnable} methods. If the augmentation method is \textit{non-learnable}, we can simply omit the parameter $\theta$ for brevity.

\begin{figure}[t]
	\centering
	\subfigure[Decoupled Training (DT)]{
	\includegraphics[width=0.99\columnwidth]{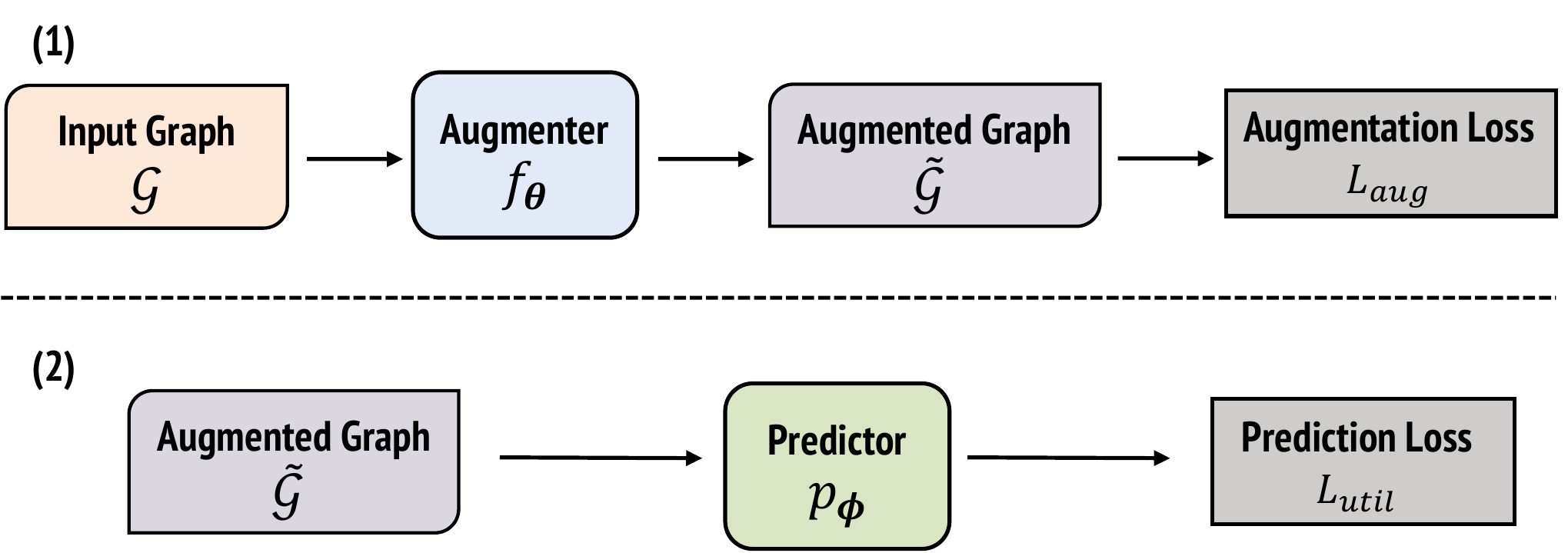}}
	
	\subfigure[Joint Training (JT)]{
	\includegraphics[width=0.99\columnwidth]{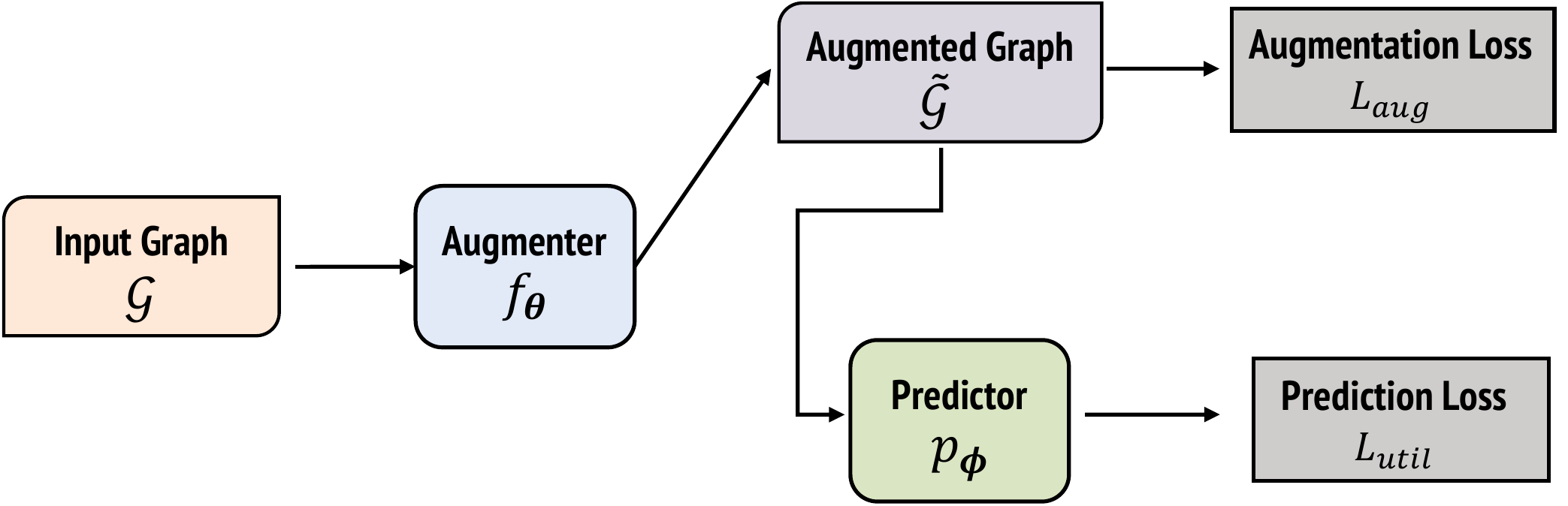}
	}
	
	\subfigure[Bi-level Optimization (BO)]{
	\includegraphics[width=0.99\columnwidth]{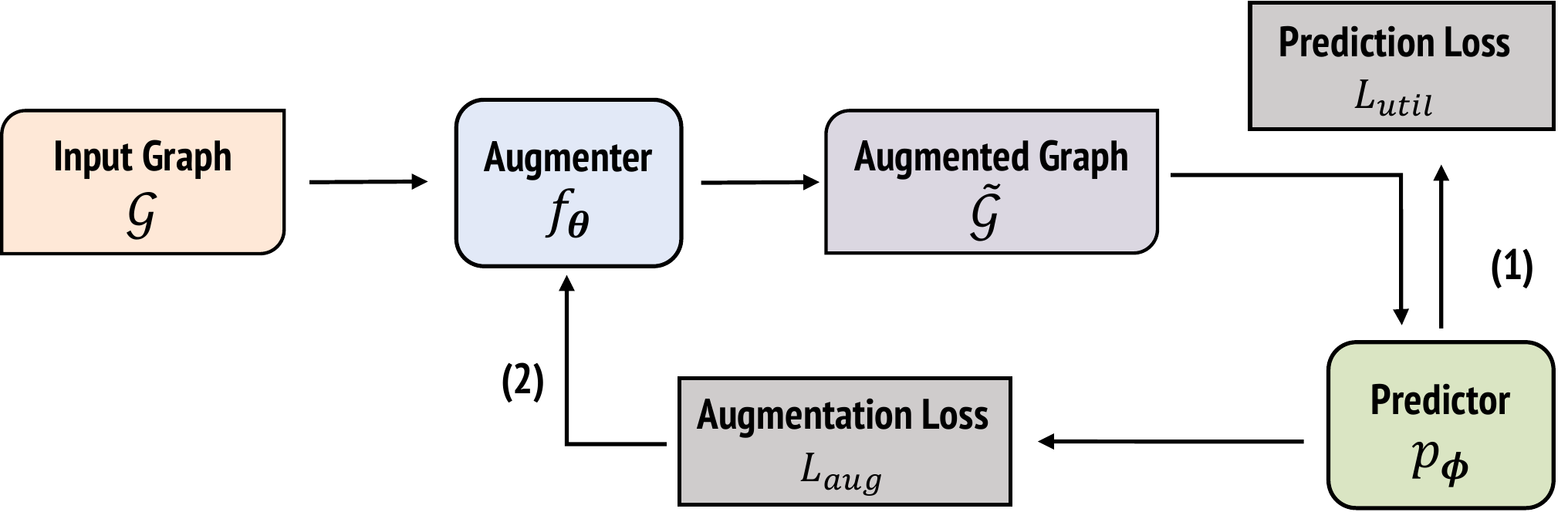}
	}
	\caption{Learnable GraphDA training paradigms.}
	\label{fig:learnable vs. non-learnable}
\end{figure}

In general, as the ultimate goal of GraphDA is to improve the GNN model performance on downstream learning tasks, we need to consider them together during the learning process. We denote the augmentation loss as $L_{\textrm{aug}}$ whose goal is to regularize the augmented graph(s) to be close to the given graph(s), and denote the utility loss that measures the GNN performance on specific downstream tasks as $L_{\textrm{utility}}$. 
In terms of training strategies, most, if not all, of the \textit{learnable} GraphDA methods can be divided into three categories: (1) decoupled training, (2) joint training, and (3) bi-level optimization. The workflow of each scheme is shown in Figure~\ref{fig:learnable vs. non-learnable} and the details can be found as follows:

\smallskip
\noindent\textbf{Decoupled Training (DT).} In this training scheme, the augmenter $f_{\theta}$ and the predictor $p_\phi$ are independently trained in a two-stage paradigm. Specifically, the augmenter is first learned with augmentation loss $L_\texttt{aug}$. After that, the prediction model $p_\theta$ is trained on the augmented graph under the supervision of specific downstream tasks (i.e., $L_{\textrm{util}}$). The learning process can be formulated as:
\begin{equation}
\begin{aligned}
      \theta^* &=  \arg \min_{\theta}\ L_\texttt{aug}\Big(\{\mathcal{G}_{i}\},\{f_{\theta}(\mathcal{G}_{i})\}\Big),\\
    \phi^* &= \arg \min_{\phi} L_\texttt{util}\Big(p_{\phi}, \{f_{\theta^*}(\mathcal{G}_{i})\}\Big).
\end{aligned}
\end{equation}


\smallskip
\noindent\textbf{Joint Training (JT).} In the joint training scheme, the augmenter $f_{\theta}$ and the predictor $p_\phi$ are jointly trained with the augmentation loss $L_{\textrm{aug}}$ and utility loss $L_{\textrm{util}}$. This learning process can be also considered as multi-task learning, which can be expressed as follows:
\begin{equation}
\begin{aligned}
    \theta^*,  \phi^*= \arg \min_{\theta, \phi}\  &L_\texttt{aug}\Big(\{\mathcal{G}_{i}\},\{f_{\theta}(\mathcal{G}_{i})\}\Big)\\
    + &L_\texttt{util}\Big(p_{\phi},\{f_{\theta}(\mathcal{G}_{i})\}\Big).
\end{aligned}
\end{equation}

\smallskip
\noindent\textbf{Bi-level Optimization (BO).} Another training scheme of GraphDA for DGL is bi-level optimization. Different from joint training, the augmenter $f_{\theta}$ and the predictor $p_\phi$ are alternatively updated with the augmentation loss $L_{\textrm{aug}}$ and utility loss $L_{\textrm{util}}$. As shown in Figure~\ref{fig:learnable vs. non-learnable} (c), the update of the augmenter is based on the optimal updated predictor, which implies a bi-level optimization problem with
$\theta$ as the upper-level variable and $\phi$ as the lower-level variable:
\begin{equation}
\begin{aligned}
      \theta^* &= \arg \min_{\theta}\ L_\texttt{aug}\Big( \{\mathcal{G}_{i}\}, \{f_{\theta}(\mathcal{G}_{i})\},  p_{\phi^*(\theta)}\Big),\\
    s.t. \quad \phi^*(\theta) &=  \arg \min_{\phi} L_\texttt{util}\Big(p_{\phi}, \{f_{\theta}(\mathcal{G}_{i})\}\Big).
\end{aligned}
\end{equation}

\begin{figure*}[t]
	\centering
	\includegraphics[width=\textwidth]{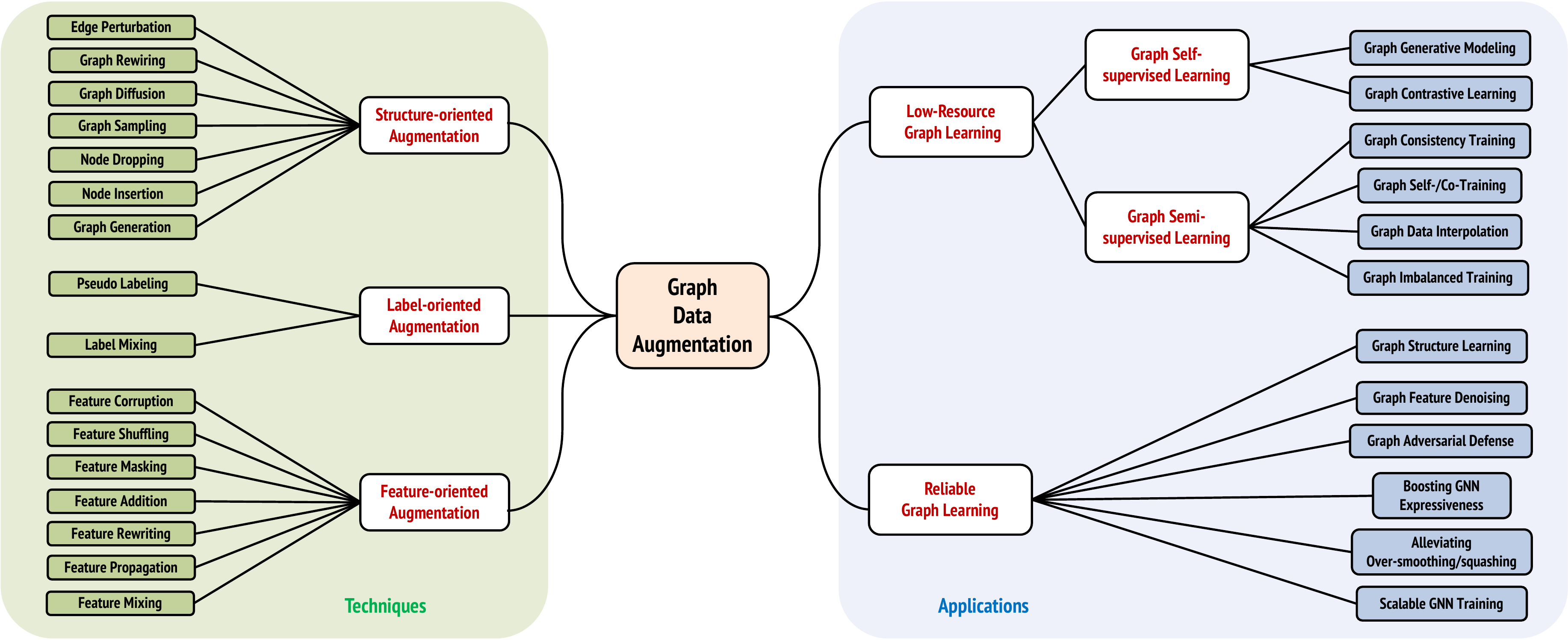}
	\caption{Proposed taxonomy of Graph Data Augmentation (GraphDA) techniques and applications.}
	\label{fig:Taxonomy}
\end{figure*}

In the following subsections, we provide a systematic taxonomy to cover mainstream GraphDA techniques. Since graphs commonly consist of multiple information modalities, GraphDA techniques can be naturally divided into three categories based on the augmentation modality, including: \emph{feature-oriented}, \emph{structure-oriented}, and \emph{label-oriented} techniques. Specifically, we summarize the commonly used techniques in terms of each augmentation modality and clearly illustrate their augmentation strategies. Figure~\ref{fig:Taxonomy} (left) summarizes our proposed taxonomy for GraphDA techniques.

\subsection{Structure-oriented Augmentations}
Different from i.i.d. data, graphs are inherently relational where the connections (i.e., edges) between data instances (i.e., nodes) are unique and essential for understanding and analyzing graphs. Given an input graph $\mathcal{G} = (\mathbf{A}, \mathbf{X})$, a structure-oriented GraphDA operation focuses on augmenting the adjacency matrix $\mathbf{A}$ of the input graph. We summarize the representative ones as follows.






\smallskip
\noindent\textbf{Edge Perturbation.} Perturbing the given graph structure, e.g., randomly adding or dropping edges, is a widely adopted GraphDA method in different DGL tasks~\cite{velickovic2019deep, you2020graph,DBLP:conf/icml/YouCSW21,zhu2021graph}. Mathematically \textit{edge perturbation} keeps the original node order and rewrites a part of the entries in the given adjacency matrices, which can be defined as follows:
\begin{equation}
\label{eq:perturb}
   \mathbf{\tilde{A}} = \mathbf{A} \oplus \mathbf{C},
\end{equation}
where $\mathbf{C}$ is the corruption matrix and $\oplus$ denotes the XOR (exclusive OR) operation. Commonly, the corruption matrix $\mathbf{C}$ is obtained by sampling, i.i.d., from a prior distribution, and $\mathbf{C}_{ij}$ determines whether to corrupt the adjacency matrix at position (i, j). For example, assuming a given corruption rate $\rho$, we may define the corruption matrix as $\mathbf{C}_{ij} \sim Bernoulli(\rho)$, whose elements in $\mathbf{C}$ are set to 1 individually with a probability $\rho$ and 0 with a probability $1 -\rho$.

%





 
\smallskip
\noindent\textbf{Graph Rewiring.} Though sharing the same basic operation with \textit{edge perturbation}, \textit{graph rewiring} has an opposite augmentation objective, which is to improve the utility of the input graph by rewiring the edges. Instead of perturbing the input graph structure by randomly adding/dropping edges, \textit{graph rewiring} is commonly guided by the learning objective of the downstream task, and the corruption matrix $\mathbf{C}$ is learned or predicted through a specific module. 


 
\smallskip
\noindent\textbf{Graph Diffusion.} As another effective structure-wise augmentation strategy for improving the graph utility, \textit{graph diffusion} generates an augmented graph by exploiting the global structure knowledge of the input graph. In certain cases, it is also considered as a \textit{graph rewiring} method~\cite{topping2022understanding}. Specifically, graph diffusion injects the global topological information into the given graph structure by connecting nodes with their indirectly connected neighbors with calculated weights. A generalized graph diffusion operation can be formulated as:
\begin{equation}
\mathbf{\tilde{A}} = \sum^{\infty}_{k = 0}\gamma_{k}\mathbf{T}^{k},
\label{eq:diff_matrix}
\end{equation}
where $\mathbf{T} \in \mathbb{R}^{N\times N}$ is the generalized transition matrix derived from the adjacency matrix $\mathbf{A}$ and $\gamma_k$ is the weighting coefficient that determines the ratio of global-local information. Imposing $\sum_{k=0}^{\infty} \gamma_k = 1, \gamma_k \in [0, 1]$ and $\lambda_i \in [0, 1]$ where
$\lambda_i$ are eigenvalues of $\mathbf{T}$, guarantees convergence. 
Two popular examples of graph diffusion are personalized PageRank (PPR)~\cite{page1999pagerank} (i.e., $\gamma_k = \alpha (1- \alpha)^k$) and
the heat kernel~\cite{kondor2002diffusion} (i.e., $\gamma_k = e^{-t} \frac{t^k}{k!}$). where $\alpha$ denotes teleport probability in a random walk and $t$ is diffusion time. Closed-form solutions to heat kernel and PPR diffusion are formulated in Eq. (\ref{eq:heat}) and (\ref{eq:PPR}), respectively:
\begin{align}
    \label{eq:heat}
    \mathbf{\tilde{A}}^{\text{heat}} &= e^{- (\mathbf{I} - \mathbf{T})t},\\
    \label{eq:PPR}
    \mathbf{\tilde{A}}^{\text{PPR}} &= \alpha (\mathbf{I}  - (1-\alpha)\mathbf{T})^{-1}.
\end{align}

\smallskip
\noindent\textbf{Graph Sampling.}
Graph sampling or subgraph sampling is a commonly used data augmentation technique for graphs. It can be used for different purposes, such as scaling up GNNs~\cite{hamilton2017inductive}, and creating augmented views~\cite{qiu2020gcc}, to name a few. The augmented graph is obtained via a sampler $\textsc{Sample}(\mathcal{G})$ which can be vertex-based sampling~\cite{jiao2020sub}, edge-based sampling~\cite{zheng2020robust}, traversal-based sampling~\cite{qiu2020gcc}, and other advanced methods such as Metropolis-Hastings sampling~\cite{park2021metropolis}. For all the above graph samplers, they commonly return a connected subgraph induced from the sampled nodes. Mathematically, graph sampling can be represented as:
\begin{equation}
    \tilde{\mathcal{G}}=\{\tilde{\mathbf{A}},\tilde{\mathbf{X}}\}=\{\mathbf{A}[idx,idx],\mathbf{X}[idx,:]\},
\end{equation}
\noindent where $idx$ is a list of index to select the given elements (i.e., rows and columns) from $\mathbf{A}$ and $\mathbf{X}$.
In general, the goal of graph sampling is to find augmented graph instances from the input graphs that best preserve desired properties by keeping a portion of nodes and their underlying linkages. For example, in the \emph{graph sparsification}~\cite{zheng2020robust,luo2021learning} problem, researchers aim to sample the subgraph which preserves as much task-relevant information as possible.

\smallskip
\noindent\textbf{Node Dropping.} In the literature, node dropping is also known as node masking. Specifically, a set of nodes $\hat{\mathcal{V}} = \{v_{i} \in \mathcal{V} \}$ will be dropped from the input graph, together with their associated edges $\hat{\mathcal{E}} = \{e_{i} \in \mathcal{E} \}$. This augmentation can be formulated as:
\begin{equation}
     \mathbf{\tilde{A}} =  \{\mathcal{V} \setminus \hat{\mathcal{V}}, \mathcal{E} \setminus \hat{\mathcal{E}}\}.
\end{equation}
Note that for attributed graphs, the corresponding node features will also be dropped at the same time.

\smallskip
\noindent\textbf{Node Insertion.} Node insertion is commonly used to improve the message-passing or connectivity on the input graph by inserting virtual node(s)~\cite{gilmer2017neural}. Specifically, node insertion adds an extra set of nodes $\hat{\mathcal{V}} = \{v_{i}\}$ and a set of edges $\hat{\mathcal{E}} = \{e_{i}\}$ between $\hat{\mathcal{V}}$ and $\mathcal{V}$ to the original node set $\mathcal{V}$ and the edge set $\mathcal{E}$, respectively:
\begin{equation}
   \mathbf{\tilde{A}} =  \{\mathcal{V} \cup \hat{\mathcal{V}}, \mathcal{E} \cup \hat{\mathcal{E}}\}.
\end{equation}
Since node insertion also requires adding additional edges in the new graph, this GraphDA operation is highly related to graph rewiring. Note that for attributed graphs, the corresponding node features also need to be initialized, e.g., using the mean average of all the connected node features.



\smallskip
\noindent\textbf{Graph Generation.} Graph generation is commonly used as a GraphDA strategy for improving the scales of training graphs graph-level DGL tasks, e.g., graph classification. Most graph generation methods are expected to automatically learn from observed graphs. Generally, the graph generation process can be presented as follows:
\begin{equation}
    \tilde{\mathcal{G}}\sim D_{\theta}(\mathcal{G}|\{\mathcal{G}_i\}),
\end{equation}
where $\{\mathcal{G}_i\}$ is the set of observed graphs and here the augmentation function $D_{\theta}$ is a graph generation distribution parameterized by $\theta$.
Notice that some techniques such as \emph{graph coarsening}~\cite{cai2021graph} and \emph{graph condensation}~\cite{jin2022graph} whose goals are to generate a new graph from the initial large graph can also be categorized into this augmentation operation. We also consider edge mixing between two (or mode) graphs~\cite{guo2021intrusion} as one instantiation of graph generation.

\subsection{Feature-oriented Augmentations}



In this subsection, we review the feature-oriented GraphDA techniques. Generally, given an input graph $\mathcal{G} = (\mathbf{A}, \mathbf{X})$, a feature-oriented GraphDA operation focuses on performing transformation on the node feature matrix $\mathbf{X}$. Notably, we also consider those methods performing augmentations on the latent feature representations $\mathbf{H}$ as feature-oriented augmentation methods.






\smallskip
\noindent\textbf{Feature Corruption.} This GraphDA method aims at adding noises to either the original node features~\cite{feng2019graph} or learned feature representations~\cite{yang2021graph}.
For simplicity, here we use $\mathbf{x}_i$ to represent the original node features or learned feature representations:
\begin{equation}
   \mathbf{\tilde{x}}_i = \mathbf{x}_i + \mathbf{r}_i,
\end{equation}
where $\mathbf{r}_i$ denotes the added feature noise. Note that the feature noise could be either randomly added~\cite{velickovic2019deep} or learned in an adversarial training fashion~\cite{feng2019graph,yang2021graph}.

\smallskip
\noindent\textbf{Feature Shuffling.}  By randomly changing the contextual information through switching rows and columns in the feature matrix, The input feature matrix $\mathbf{X}$ is corrupted to yield augmentations. This method can be formulated as:
\begin{equation}
\mathbf{\tilde{X}} =  \mathbf{P}_r \mathbf{X} \mathbf{P}_c,
\end{equation}
where $\mathbf{P}_r$ and $\mathbf{P}_c$ are row-wise permutation matrix and column-wise permutation matrix, respectively.  have exactly one entry of $1$ in each row and each column and $0$ elsewhere.

\smallskip
\noindent\textbf{Feature Masking.} The core operation of feature masking is to set a part of the entries in the node feature matrix $\mathbf{X}$ to $0$, which can be formulated as:
\begin{equation}
    \mathbf{\tilde{X}} =   \mathbf{X} \odot  \mathbf{M}
\end{equation}
where $\mathbf{M}$ is the masking matrix that $\mathbf{M}_{i,j} = 0$ if the $j$-th element of vector $i$ is masked/dropped, otherwise $\mathbf{M}_{i,j} = 1$. The masking matrix $\mathbf{M}$ is commonly generated by the Bernoulli distribution~\cite{you2020graph,DBLP:conf/icml/YouCSW21,thakoor2021large}.




\smallskip
\noindent\textbf{Feature Addition.} Since the input graph usually lacks informative features in real-world scenarios, Feature Addition can be used to (1) initiate node features on plain graphs to smoothly incorporate into DGL models (e.g., GNNs) and (2) supplement additional graph knowledge that are hard to be captured by GNN models. A straightforward way is to encode proximate/topological information (e.g., node index or node properties) into a feature vector and concatenate with the original node features. In general, Feature Addition can be expressed as:
\begin{equation}
    \tilde{\mathbf{x}}_i =  [\hat{\mathbf{x}}_i || \mathbf{x}_i],
\end{equation}
where $||$ denotes the concatenation operation and $\mathbf{x}_i$ could be an empty vector if the input graph is plain.


\smallskip
\noindent\textbf{Feature Rewriting.} Considering the fact that the given node features are commonly noisy and incomplete, recovering the clean and complete node features can directly improve the performance of the DGL models. Generally, feature rewriting can be expressed as:
\begin{equation}
         \tilde{\mathbf{x}} =  \alpha \mathbf{x}_i + \beta\mathbf{b}_i,
\end{equation}
where $\alpha$ and $\beta$ are two controlling parameters and $\mathbf{b}_i$ is a feature vector computed in a heuristic or learnable way. For example, Wang et al. propose \textit{feature replacement}~\cite{wang2020nodeaug} that rewrites node features with its neighbors' features. While Xu et al.~\cite{xu2021graph} apply gradient descent-based optimizers to rewrite the node features as parameters.

\smallskip
\noindent\textbf{Feature Mixing.} Based on the features of nodes in the input graph, feature mixing can be used to obtain the node features of a synthetic node:
\begin{equation}
         \tilde{\mathbf{x}}  = \lambda \mathbf{x}_i + (1 - \lambda)\mathbf{x}_j,
\end{equation}
where $\lambda$ is the mixing coefficient that controls the proportion of information from $\mathbf{x}_i$ and $\mathbf{x}_j$. Notably, feature mixing can also be performed on the intermediate representations learned from two training samples (i.e., Manifold Mixup~\cite{verma2019manifold}).



\smallskip
\noindent\textbf{Feature Propagation.} Based on Graph Diffusion, Feature Propagation propagates the node features along the graph structure. It is an interpolation method that has also been widely used to augment the node features of the input graph. Mathematically:
\begin{equation}
    \mathbf{\tilde{X}}  = \mathbf{\tilde{A}} \mathbf{X},
\end{equation}
where $\mathbf{\tilde{A}}$ is the new adjacency matrix obtained through different graph diffusion methods.

\subsection{Label-oriented Augmentations}
\label{sec:label-wise augmentation}
Due to the expensive data labeling cost on graphs, label-oriented GraphDA is an important line of work to directly enrich the limited labeled training data. Commonly, there are two groups of strategies.


\smallskip
\noindent\textbf{Pseudo-Labeling.} Pseudo-labeling is a semi-supervised learning mechanism that aims to obtain one (or several) augmented labeled set(s), based on their most confident predictions on unlabeled data. Its learning process starts with a base teacher model trained on the labeled set $\mathcal{D}^L$, and then the teacher model is applied to the unlabeled data $\mathcal{D}^U$ to obtain pseudo labels (hard or soft) of unlabeled data. Finally, a subset of unlabeled data $\mathcal{D}^P$ will be used to augment the training data, and the combined data $\mathcal{D}^L \cup \mathcal{D}^P$ can be used to train a student model. In this sense, the label signals can be ``propagated'' to the unlabeled data samples via the learned teacher model. Note that this learning process could go through multiple rounds until converges by iteratively updating the teacher model with the current student model.






\smallskip
\noindent\textbf{Label Mixing.} In order to enlarge the scale of training samples, we can directly interpolate the training samples based on the labeled examples. Generally, Mixup~\cite{zhang2018mixup} constructs virtual training samples via feature mixing and label mixing:
\begin{equation}
\begin{aligned}
     \tilde{\mathbf{x}} &= \lambda \mathbf{x}_i + (1 - \lambda)\mathbf{x}_j,\\
    \tilde{\mathbf{y}} &= \lambda \mathbf{y}_i + (1 - \lambda)\mathbf{y}_j,
\end{aligned}
\end{equation}
where $(\mathbf{x}_i, \mathbf{y}_i)$ and $(\mathbf{x}_j, \mathbf{x}_j)$ are two labeled samples randomly sampled from the training set, and $\lambda \in [0, 1]$. In this way, Mixup methods extend the training distribution by incorporating the prior knowledge that linear interpolations of feature vectors should lead to linear interpolations of the associated target labels.

\begin{table*}[t]
\centering
\caption{Summary of representative GraphDA works for graph \textit{self-supervised} learning.  DT, JT, BO stand for decoupled training, joint training, and bi-level optimization, respectively}
\label{tab:summary of graph self-supervised learning}
\scalebox{0.7}{
\begin{tabular}{ccccccccccc}
\toprule
\multirow{2}{*}{Topic} & \multirow{2}{*}{Name} & \multirow{2}{*}{Ref.} & \multirow{2}{*}{Year} & \multirow{2}{*}{Venue} & \multirow{2}{*}{Task Level} & \multicolumn{3}{c}{Augmented Data Modality} & \multirow{2}{*}{Learnable} &
\multirow{2}{*}{Augmentation Technique}\\
&  &  &  &  &  & Structure & Feature & Label \\\midrule
\multirow{10}{*}{\makecell{Graph Generative\\Modeling}}& GraphMAE & \cite{hou2022graphmae} & 2022 & KDD & node \& graph & \checkmark & & & & feature masking\\\cmidrule{2-11}
& GMAE & \cite{chen2022graph} & 2022 & arXiv & noded \& graph & \checkmark & & & & node dropping\\\cmidrule{2-11}
& MGAE & \cite{tan2022mgae} & 2022 & arXiv & node \& edge & \checkmark & & & & edge perturbation (dropping) \\\cmidrule{2-11}
& MGM & \cite{mahmood2021masked} & 2021 & \makecell{Nature\\ Communications} & graph & & \checkmark & & & feature corruption\\\cmidrule{2-11}
 & GPT-GNN & \cite{hu2020gpt} & 2020 & KDD & node & \checkmark & \checkmark &  &  & \makecell{edge perturbation (dropping)\\feature masking}\\\cmidrule{2-11}
& MTL & \cite{you2020does} & 2020 & ICML & node &  & \checkmark & \checkmark & \checkmark(JT) & \makecell{pseudo-labeling/feature corruption}\\\cmidrule{2-11}
& GraphBert & \cite{zhang2020graph} & 2020 & arXiv & node & \checkmark & \checkmark &  &  & \makecell{graph sampling}\\\cmidrule{2-11}

& Pre-train & \cite{hu2019pre} & 2019 & arXiv & node \& edge \& graph & \checkmark &  & \checkmark &  & \makecell{edge perturbation/pseudo-labeling}\\\midrule

\multirow{33}{*}{\makecell{Graph Contrastive\\Learning}} & S$^3$-CL & \cite{ding2022structural} & 2022 & arXiv & node &  & \checkmark & \checkmark & \checkmark(JT) & \makecell{feature propagation/pseudo labeling}\\\cmidrule{2-11}
& SUGRL & \cite{mo2022simple} & 2022 & AAAI & node & \checkmark & \checkmark &  &  & \makecell{feature shuffling/graph sampling}\\\cmidrule{2-11}
& LG2AR & \cite{hassani2022learning} & 2022 & arXiv & node \& graph & \checkmark & \checkmark & & \checkmark(BO) & \makecell{node dropping/edge perturbation \\graph sampling/feature corruption}\\\cmidrule{2-11}
& BGRL & \cite{thakoor2021large} & 2022 & ICLR & node & \checkmark & \checkmark &  &  & \makecell{edge perturbation (dropping)\\feature masking}\\\cmidrule{2-11}
& ARIEL & \cite{feng2022adversarial} & 2022 & TheWebConf & node & \checkmark & \checkmark &  & \checkmark(JT) & \makecell{feature corruption/edge perturbation}\\\cmidrule{2-11}
& AD-GCL & \cite{suresh2021adversarial} & 2021 & NeurIPS & graph & \checkmark &  &  & \checkmark(JT) & edge perturbation (dropping)\\\cmidrule{2-11}
& JOAO & \cite{DBLP:conf/icml/YouCSW21} & 2021 & ICML & graph & \checkmark & \checkmark & & \checkmark(BO) & \makecell{node dropping/edge perturbation\\graph sampling/feature corruption}  \\\cmidrule{2-11}
& GCA & \cite{zhu2021graph} & 2021 & TheWebConf & node & \checkmark & \checkmark &  &  & \makecell{edge perturbation (dropping)\\feature masking}\\\cmidrule{2-11}
& MERIT & \cite{jin2021multi} & 2021 & IJCAI & graph & \checkmark & \checkmark &  &  & \makecell{graph diffusion/edge perturbation\\graph sampling/feature masking}\\\cmidrule{2-11}
& CSSL & \cite{zeng2021contrastive} & 2021 & AAAI & graph & \checkmark &  &  &  & \makecell{node dropping/node insertion\\edge perturbation}\\\cmidrule{2-11}

& GraphCL & \cite{you2020graph} & 2020 & NeurIPS & graph & \checkmark & \checkmark &  &  & \makecell{node dropping/edge perturbation\\graph sampling/feature corruption}\\\cmidrule{2-11}
& GCC & \cite{qiu2020gcc} & 2020 & KDD & graph & \checkmark &  &  &  & graph sampling\\\cmidrule{2-11}

& SUBGCON & \cite{jiao2020sub} & 2020 & ICDM & node & \checkmark & & & & graph sampling \\\cmidrule{2-11}
& MVGRL & \cite{hassani2020contrastive} & 2020 & ICML & node \& graph & \checkmark & & & & \makecell{graph diffusion/graph sampling} \\\cmidrule{2-11}
& GRACE & \cite{zhu2020deep} & 2020 & arXiv & node & \checkmark & \checkmark & & & \makecell{edge perturbation (dropping)\\feature masking} \\\cmidrule{2-11}

& DGI & \cite{velickovic2019deep} & 2019 & ICLR & node & & \checkmark & & & feature corruption\\
\bottomrule

\end{tabular}
}
\end{table*}

\section{Graph Data Augmentation for \\Low-Resource Graph Learning}
\label{sec: low-resource application}
The shortage of ground-truth labels has been a longstanding and notorious problem for learning effective DGL. To advance the research on \textit{Low-resource Graph Learning}, GraphDA has been actively investigated and shows promising results. In this section, we discuss the applications of GraphDA for solving both \textit{Graph Self-Supervised Learning} and \textit{Graph Semi-Supervised Learning}. We summarize the representative works in Table \ref{tab:summary of graph self-supervised learning} and Table \ref{tab:summary of graph semi-supervised learning}, respectively.


\begin{figure}[t]
	\centering
	\subfigure[Graph Generative Modeling]{
	\includegraphics[width=\columnwidth]{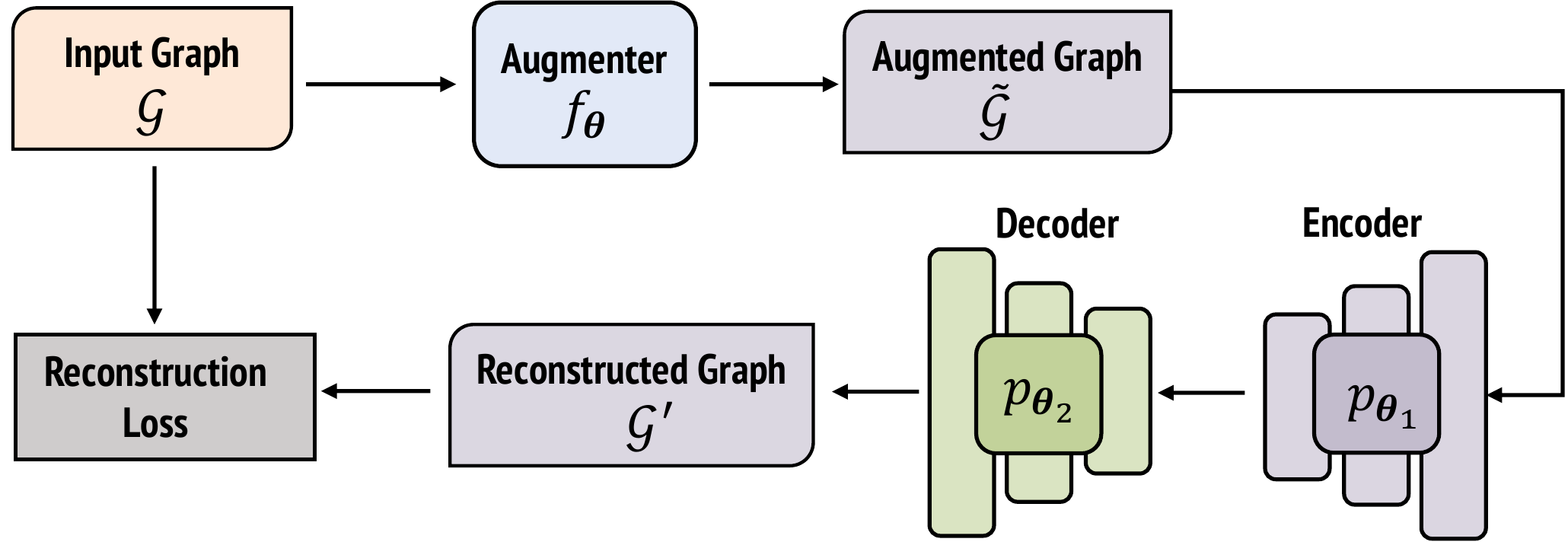}
	}
	
	\subfigure[Graph Contrastive Learning]{
	\includegraphics[width=\columnwidth]{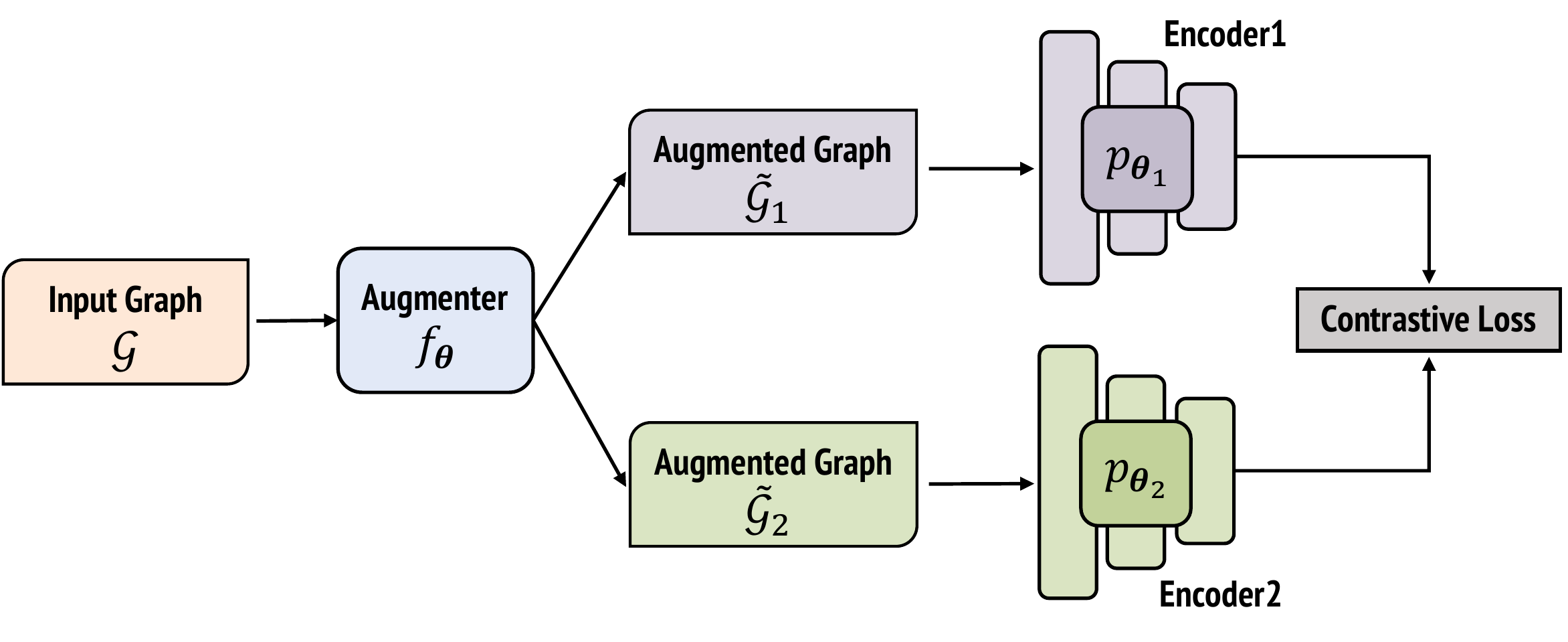}
	}
	\caption{Comparison between the workflows of Graph Generative Modeling and Graph Contrastive Learning.}%
	\label{fig:selfcompare}
\end{figure}

	


\subsection{Graph Self-Supervised Learning}
\smallskip
\noindent\textbf{Graph Generative Modeling.} Recently data augmentation has been widely used for graph self-supervised learning (SSL). Based on the idea of Graph AutoEncoders (GAE)~\cite{kipf2016variational,ding2019deep}, graph generative modeling methods perform data augmentation on the input graphs through either \textbf{\textit{edge perturbation}}, \textbf{\textit{feature masking}}, or \textbf{\textit{node dropping (masking)}} then learn the node representations by reconstructing feature or/and structure information from the augmented graphs (as shown in Fig. \ref{fig:selfcompare} (a)). For example, Denoising Link Reconstruction~\cite{hu2019pre} randomly drops existing edges to obtain the perturbed graph and tries to recover the discarded connections with a pairwise similarity-based decoder trained by the cross-entropy loss. Given a graph with its nodes and edges randomly masked, GPT-GNN~\cite{hu2020gpt} generates one masked node and its related edges jointly and optimizes the likelihood of the node and edges generated in the current iteration.  GraphBert~\cite{zhang2020graph} samples linkless subgraphs and pre-trains a graph transformer model via node feature reconstruction and graph structure recovery. You et al.~\cite{you2020does} define the graph completion pretext task which aims to recover the masked feature of target nodes based on their neighbors’ features and connections. MGM~\cite{mahmood2021masked} tries to reconstruct the masked node/edge features for learning the GNNs for molecule generation. GMAE~\cite{chen2022graph} and GraphMAE~\cite{hou2022graphmae} first randomly mask the features of some nodes and then their decoders reconstruct the original features of the masked nodes, while MGAE~\cite{tan2022mgae} tries to reconstruct the masked edges based on the augmented graphs.


\smallskip
\noindent\textbf{Graph Contrastive Learning.} Motivated by the recent breakthroughs in contrastive visual feature learning, data augmentation has also been widely used for Graph Contrastive Learning (GCL). As illustrated in Fig. \ref{fig:selfcompare} (b), GCL methods try to generate augmented examples from the input and view two augmented examples from the same original sample as a positive pair, while those from different original samples are negative pairs. By applying contrastive learning loss, the positive pairs will be pulled together and the negative pairs will be pushed away in the latent space. Therefore, GraphDA plays an essential role in GCL. As a pioneering work, DGI~\cite{velickovic2019deep} applies \textbf{\textit{feature shuffling}} and \textbf{\textit{edge perturbation}} to obtain the negative pairs of graphs, then a contrastive objective is proposed to maximize the mutual information between node embeddings and a global summary embedding. Another popular GraphDA method for GCL is \textbf{\textit{graph sampling}}. As an example, GCC~\cite{qiu2020gcc} proposes to sample subgraphs as contrastive instances to pre-train the graph encoder, which can be used for different downstream tasks with either freezing or full fine-tuning strategy. SUBGCON~\cite{jiao2020sub} also utilizes a subgraph sampler based on Personalized PageRank to sample the augmented subgraph and perform contrastive learning between the representations of the central node and the sampled subgraph.


%

It is noteworthy that many GCL methods usually leverage combinations of augmentation strategies. For instance, MVGRL~\cite{hassani2020contrastive} first augments the input graph via \textbf{\textit{graph diffusion}}. Afterward, two graph views are generated by \textbf{\textit{subgraph sampling}} and the model learns to contrast node representations to global representations across the two views. GRACE~\cite{zhu2020deep} adopts two augmentation strategies, i.e., \textbf{\textit{edge perturbation (dropping)}} and \textbf{\textit{feature masking}}, to generate augmented views of graph data. It jointly considers both intra-view and inter-view negative pairs for the contrastive learning objectives. gCooL~\cite{li2022graph} uses the same augmentations and considers the community information in GCL. GraphCL~\cite{you2020graph} considers four GraphDA operations: \textbf{\textit{node dropping}}, \textbf{\textit{edge perturbation}}, \textbf{\textit{feature masking}}, and \textbf{\textit{subgraph sampling}}, while MERIT~\cite{jin2021multi} leverages \textbf{\textit{graph diffusion}}, \textbf{\textit{edge perturbation (dropping)}}, \textbf{\textit{subgraph sampling}}, and \textbf{\textit{feature masking}} to generate augmented graphs. CSSL~\cite{zeng2021contrastive} augments graphs with the \textbf{\textit{edge perturbation (dropping)}} and \textbf{\textit{node dropping}}. Recent work SUGRL~\cite{mo2022simple} leverages \textbf{\textit{feature shuffling}} and \textbf{\textit{graph sampling}}, S$^3$-CL uses \textbf{\textit{feature propagation}} and \textbf{\textit{pseudo labeling}} to efficiently perform GCL.

Different from the aforementioned pre-defined GraphDA strategies, another line of research proposes to perform data augmentation on graphs in a learnable manner. Specifically, Zhu et al.~\cite{zhu2021graph} propose a joint, adaptive data augmentation scheme based on \textbf{\textit{edge perturbation (dropping)}} and \textbf{\textit{feature masking}} to provide diverse contexts for nodes in different views, so as to boost optimization of the contrastive objective. GASSL~\cite{yang2021graph} is an adversarial self-supervised learning framework for learning unsupervised representations of graph data without any handcrafted views. It learns to perform \textbf{\textit{feature corruption}} on either the input or latent space. AD-GCL~\cite{suresh2021adversarial} enables GNNs to avoid capturing redundant information during the training by optimizing \textbf{\textit{edge perturbation (dropping)}} strategy used in GCL in an adversarial fashion. To automatically select optimal augmentation combinations for the given graph dataset, JOAO~\cite{DBLP:conf/icml/YouCSW21} and LG2AR~\cite{hassani2022learning} are proposed to automatically select augmentations from a given pool of augmentation types: \{\textbf{\textit{node dropping}}, \textbf{\textit{subgraph sampling}}, \textbf{\textit{edge perturbation}}, \textbf{\textit{feature masking}}\}. It is worth mentioning that there are increasingly more works in GCL, we will not cover all of them due to the space limit.







\begin{table*}[t]
\centering
\caption{Summary of representative GraphDA works for graph \textit{semi-supervised} learning.  DT, JT, BO stand for decoupled training, joint training, and bi-level optimization, respectively}
\label{tab:summary of graph semi-supervised learning}
\scalebox{0.7}{
\begin{tabular}{ccccccccccc}
\toprule
\multirow{2}{*}{Topic} & \multirow{2}{*}{Name} & \multirow{2}{*}{Ref.} & \multirow{2}{*}{Year} & \multirow{2}{*}{Venue} & \multirow{2}{*}{Task Level} & \multicolumn{3}{c}{Augmented Data Modality} & \multirow{2}{*}{Learnable} &
\multirow{2}{*}{Augmentation Technique}\\
&  &  &  &  &  & Structure & Feature & Label \\
\midrule
\multirow{6}{*}{\makecell{Graph \\Consistency Training}} & NodeAug & \cite{wang2020nodeaug} & 2020 & KDD & node & \checkmark & \checkmark & \checkmark &  & \makecell{feature rewriting/graph rewiring\\pseudo labeling}\\\cmidrule{2-11}
& GRAND & \cite{feng2020graph} & 2020 & NeurIPS & node &  & \checkmark & \checkmark &  & \makecell{node dropping/feature propagation\\pseudo labeling}\\\cmidrule{2-11}
& SCR & \cite{zhang2021improving} & 2021 & arXiv & node & \checkmark &  &  &  & graph rewiring \\\cmidrule{2-11}
& MH-Aug & \cite{park2021metropolis} & 2021 & NeurIPS & node & \checkmark &  & \checkmark &  & graph sampling\\\cmidrule{2-11}
& NASA & \cite{bo2022regularizing} & 2022 & AAAI & node & \checkmark &  & \checkmark &  & edge perturbation (dropping)\\\midrule

\multirow{10}{*}{\makecell{Graph \\Self-/Co-Training}} & Meta-PN & \cite{ding2022meta} & 2022 & AAAI & node &  &  & \checkmark & \checkmark(BO) & pseudo-labeling\\\cmidrule{2-11}
& NRGNN & \cite{dai2021nrgnn} & 2021 & KDD & node & \checkmark &  & \checkmark & \checkmark(DT) & \makecell{graph rewiring/pseudo-labeling} \\\cmidrule{2-11}
& PTA & \cite{dong2021equivalence} & 2021 & TheWebConf & node &  &  & \checkmark &  & pseudo-labeling\\\cmidrule{2-11}
& CGCN & \cite{hui2020collaborative} & 2020 & AAAI & node &  &  & \checkmark & \checkmark(JT) & pseudo-labeling\\\cmidrule{2-11}
& M3S & \cite{sun2020multi} & 2020 & AAAI & node &  &  & \checkmark & \checkmark(JT) & pseudo-labeling\\\cmidrule{2-11}
& Co-GCN & \cite{li2020co} & 2020 & AAAI & node & \checkmark &  &  & \checkmark(DT) & \makecell{feature masking/pseudo labeling}\\\cmidrule{2-11}
& ST-GCNs & \cite{li2018deeper} & 2018 & AAAI & node &  &  & \checkmark & \checkmark(DT) & pseudo-labeling\\\midrule

\multirow{6}{*}{\makecell{Graph \\Data Interpolation}} & G-Transplant & \cite{park2022graph} & 2022 & AAAI & graph & \checkmark & \checkmark & \checkmark & \checkmark(DT) & \makecell{label mixing/graph rewiring} \\\cmidrule{2-11}
& G-Mixup & \cite{wang2021mixup} & 2021 & TheWebConf & graph &  & \checkmark & \checkmark & \checkmark(DT) & \makecell{graph generation/label mixing}\\\cmidrule{2-11}
& GraphMix & \cite{verma2021graphmix} & 2021 & AAAI & node &  & \checkmark & \checkmark &  & feature mixing/label mixing/pseudo labeling\\\cmidrule{2-11}
& ifMixup & \cite{guo2021intrusion} & 2021 & arXiv & graph & \checkmark & \checkmark & \checkmark &  & \makecell{feature mixing/label mixing/graph generation} \\\midrule

\multirow{8}{*}{\makecell{Graph \\Imbalanced Training}} & GATSMOTE & \cite{liu2022gatsmote} & 2022 & Mathematics & node & \checkmark & \checkmark & \checkmark & \checkmark(JT) & feature rewriting/node insertion/graph rewiring\\\cmidrule{2-11}
& GNN-CL & \cite{li2022graph} & 2022 & SDM & node & \checkmark & \checkmark & \checkmark & \checkmark(JT) & feature mixing/node insertion/graph rewiring\\\cmidrule{2-11}
& GraphSMOTE & \cite{zhao2021graphsmote} & 2021 & WSDM & node & \checkmark & \checkmark & \checkmark & \checkmark(JT) & feature mixing/node insertion/graph rewiring\\\cmidrule{2-11}
& DPGNN & \cite{wang2021distance}& 2021 & arXiv & node & & & \checkmark & & pseudo-labelling\\\cmidrule{2-11}
& D-GCN & \cite{wang2021dual} & 2021 & ICCSAE & node & \checkmark & & & & graph rewiring\\\cmidrule{2-11}
& GraphMixup & \cite{wu2021graphmixup} & 2021 & arXiv & node & \checkmark & \checkmark & \checkmark & \checkmark(BO) & \makecell{feature mixing/label mixing/graph rewiring}\\\bottomrule
\end{tabular}}
\end{table*}

\subsection{Graph Semi-Supervised Learning}

\smallskip
\noindent\textbf{Graph Consistency Training.} Similar to the idea of contrastive learning, Consistency Training~\cite{xie2020unsupervised} leverages unlabeled data to improve model performance by enforcing the consistency across predictions learned from different stochastic augmentations of the same input. As one effective semi-supervised learning paradigm, consistency training has also been explored in learning graph neural networks under low-resource settings. For example, NodeAug~\cite{wang2020nodeaug} uses local structure-wise augmentation operations (i.e., \textbf{\textit{feature corruption}}, and \textbf{\textit{edge perturbation}}), and minimizes the KL-divergence between the node representations learned from the original graph and augmented graph. GRAND~\cite{feng2020graph} creates multiple different augmented graphs with \textbf{\textit{node dropping}} and \textbf{\textit{feature masking}}, followed by \textbf{\textit{feature propagation}}. Then the consistency loss is applied to minimize the distances of the representations learned from the augmented graphs. Following GRAND, Zhang et al.~\cite{zhang2021improving} propose to use two \textbf{\textit{edge perturbation}} methods -- DropEdge~\cite{rong2019dropedge} or DropNode~\cite{feng2020graph} as the augmentation strategies and leverage a mean-teacher consistency regularization to guide the training of the GNN model by calculating a consistency loss between the student and teacher models. To enable consistency training on large-scale graphs, Hawkins et al.~\cite{hawkins2021scalable} propose to use \textbf{\textit{graph sampling}} to generate different neighborhood subgraph expansions and ensemble the predictions to provide pseudo labels. To avoid the detrimental effects of arbitrary augmentations, Park et al.~\cite{park2021metropolis} propose a \textbf{\textit{graph sampling}} method MH-Aug that uses the Metropolis-Hastings algorithm to obtain the augmented samples of the input graph, and adopt consistency training to better utilize the unlabeled data. Following the idea of \textbf{\textit{graph rewiring}}, NASA~\cite{bo2022regularizing} generates graph augmentations with high consistency and diversity by replacing immediate neighbors with remote neighbors and enforcing the predictions of augmented neighbors to be consistent. 



%

\smallskip
\noindent\textbf{Graph Self-/Co-Training.} To address the data scarcity issue, one effective solution is to leverage the unlabeled data to augment the limited labeled data. Following the idea of \textbf{\textit{pseudo labeling}}, self-training~\cite{yarowsky1995unsupervised} imputes the labels on unlabeled data based on a teacher model trained with limited labeled data, and it has become a prevailing paradigm to solve the problem of semi-supervised node classification when training data is limited. Among those methods, Li et al.~\cite{li2018deeper} first combine GCNs with self-training to expand supervision signals. CGCN~\cite{hui2020collaborative} generates pseudo labels by combining variational graph auto-encoder with Gaussian mixture models. Furthermore, M3S~\cite{sun2020multi} propose the multi-stage self-training and utilize a clustering method to eliminate the pseudo labels that may be incorrect. Similar ideas can also be found in \cite{dai2021nrgnn}. In addition, recent research~\cite{dong2021equivalence,ding2022meta} adopt label propagation as the teacher model to generate pseudo labels that encode valuable global structure knowledge.


Similar to Self-training, Co-training~\cite{blum1998combining} has also been investigated for augmenting the training set with unlabeled data. It learns two classifiers with initial labeled data on the two views respectively and lets them label unlabeled data for each other to augment the training data. Li et al.~\cite{li2020co} develop a novel multi-view semi-supervised
learning method Co-GCN based on \textbf{\textit{feature masking}}, which unifies GCN and co-training into one framework.

\smallskip
\noindent\textbf{Graph Data Interpolation.} Another way of obtaining extra training examples is to use interpolation-based data augmentation strategy, such as Mixup~\cite{zhang2018mixup} to generate synthetic training examples (i.e., \textbf{\textit{node insertion}}) based on \textbf{\textit{feature mixing}} and \textbf{\textit{label mixing}}. While unlike images or natural sentences, graphs have arbitrary structure, it remains a non-trivial task to identify the meaningful connections between the original nodes and synthetic nodes. Also, due to the cascading effect of graph data, even simply adding an edge into a graph can dramatically change the graph semantic meanings. To circumvent those challenges, Manifold Mixup~\cite{verma2019manifold} has been applied to graph data interpolation. Specifically, GraphMix~\cite{verma2021graphmix} trains a fully connected network (FCN) jointly with the GNN via parameter sharing, and the FCN is learned based on Manifold Mixup and \textbf{\textit{pseudo-labeling}}, which can effectively train GNNs for semi-supervised node classification. Similarly, Wang et al.~\cite{wang2021mixup} also follow the idea of Manifold Mixup and interpolate the input features of both nodes and graphs in the embedding space. Those methods leverage a simple way to avoid dealing with the arbitrary structure in the input space for mixing a node or graph pair, through mixing the graph representation learned from GNNs. 



 For the input-level graph data interpolation, ifMixup~\cite{guo2021intrusion} targets the Manifold Intrusion issue (mixing graph pairs may naturally create graphs with identical structure but with conflict labels) and first interpolates both the node features and the edges of the input pair based on \textbf{\textit{feature mixing}} and \textbf{\textit{graph generation}}. Graph Transplant~\cite{park2022graph} is another input-level graph interpolation method that leverages \textbf{\textit{graph rewiring}} to mix two dissimilar-structured graphs by replacing the destination subgraph with the source subgraph while preserving the local structure. $G$-Mixup~\cite{han2022g} first estimates the graphon of each class and then mixup between the graphons and perform \textbf{\textit{graph generation}} to generate interpolated graphs, which improves the generalizability and robustness of GNNs for semi-supervised graph classification.


\smallskip
\noindent\textbf{Graph Imbalanced Training.}
The class distribution of graph data is inherently imbalanced which follows the power-law distribution. As an example, on the benchmark Pubmed dataset, nodes are labeled into three classes while the minority class only contains $5.25\%$ of the total nodes. Such highly imbalanced data will lead to the suboptimal performance of downstream tasks especially classification tasks and one of the effective solutions is to augment the minority to alleviate the imbalance. To counter this problem, \textbf{\textit{node insertion}} has been proven as an effective solution to augment the minority class. Meanwhile, \textbf{\textit{feature mixing}} and \textbf{\textit{graph rewiring}} are also needed for enable graph imbalanced training. For instance, GraphSMOTE~\cite{zhao2021graphsmote} augments the minority class by mixing up the minority nodes and leverages an edge generator to predict neighbor information for those synthetic nodes. GraphMixup~\cite{wu2021graphmixup} first performs interpolation on a node from one target minority class with its nearest neighbors, then adopts an edge prediction module to predict the connections between generated nodes and existing nodes. Following this idea, GATSMOTE~\cite{liu2022gatsmote} and GNN-CL~\cite{li2022graph} adopts an attention mechanism to generate the edges between the synthetic nodes and the original nodes.  On the label level, DPGNN~\cite{wang2021distance} conducts \textbf{\textit{pseudo labeling}} via label propagation to enrich the training samples from the minority class. However, many challenges in this topic are still under-explored. For example, if the amount of labeled minority nodes is extremely small, such as few-shot or even one-shot per class, how to transfer knowledge from the majority classes to augment the minority classes is worth studying.

\begin{table*}[t!]
\centering
\caption{Summary of representative GraphDA works for \textit{reliable} graph learning. DT, JT, BO stand for decoupled training, joint training, and bi-level optimization, respectively.}
\label{tab:summary of robust graph learning}
\scalebox{0.7}{
\begin{tabular}{ccccccccccc}
\toprule
\multirow{2}{*}{Topic} & \multirow{2}{*}{Name} & \multirow{2}{*}{Ref.} & \multirow{2}{*}{Year} & \multirow{2}{*}{Venue} & \multirow{2}{*}{Task Level} & \multicolumn{3}{c}{Augmented Data Modality} & \multirow{2}{*}{Learnable} &
\multirow{2}{*}{Augmentation Technique}\\
&  &  &  &  &  & Structure & Feature & Label\\
\midrule
\multirow{15}{*}{\makecell{Graph Structure\\ Learning}} & DGM & \cite{kazi2022differentiable} & 2022 & TPAMI & node & \checkmark & \checkmark &  & \checkmark (JT) & \makecell{feature rewriting/graph rewiring}\\\cmidrule{2-11}
& GEN & \cite{wang2021graph} & 2021 & TheWebConf & node & \checkmark &  &  & \checkmark (JT) & graph rewiring\\\cmidrule{2-11}
& PTDNet & \cite{luo2021learning} & 2021 & WSDM & node \& edge & \checkmark &  &  & \checkmark (JT) & graph sampling\\\cmidrule{2-11}
& GAUG & \cite{zhao2021data} & 2021 & AAAI & node & \checkmark &  &  & \checkmark (DT) & graph rewiring\\\cmidrule{2-11}
& HGSL & \cite{zhao2021heterogeneous} & 2021 & AAAI & node & \checkmark &  &  & \checkmark (JT) & graph rewiring\\\cmidrule{2-11}
& IDGL & \cite{chen2020iterative} & 2020 & NeurIPS & node \& graph & \checkmark &  &  & \checkmark (JT) & graph rewiring\\\cmidrule{2-11}
& NeuralSparse & \cite{zheng2020robust} & 2020 & ICML & node & \checkmark &  &  & \checkmark (JT) & graph sampling\\\cmidrule{2-11}
& TO-GNN & \cite{yang2019topology} & 2019 & IJCAI & node & \checkmark &  &  & \checkmark (JT) & graph rewiring\\\cmidrule{2-11}
& LDS & \cite{franceschi2019learning} & 2019 & ICML & node & \checkmark &  &  & \checkmark (BO) & graph rewiring\\\cmidrule{2-11}
& PG-LEARN & \cite{wu2018quest} & 2018 & CIKM & node & \checkmark &  &  & \checkmark (JT) & graph rewiring\\\midrule

\multirow{8}{*}{\makecell{Graph Feature\\ Denoising}} & HGCA & \cite{he2022analyzing} & 2022 & TNNLS & node & & \checkmark & & \checkmark (JT) & feature addition\\\cmidrule{2-11}
& AirGNN & \cite{liu2021graph} & 2021 & NeurIPS & node &  & \checkmark &  & \checkmark (JT) & feature rewriting\\\cmidrule{2-11}
& GCNMF & \cite{taguchi2021graph} & 2021 & FGCS & node \& edge &  & \checkmark &  & \checkmark (JT) & feature addition\\\cmidrule{2-11}
& HGNN-AC & \cite{jin2021heterogeneous} & 2021 & TheWebConf & node & & \checkmark & & \checkmark (JT) & feature addition\\\cmidrule{2-11}
& SAT & \cite{chen2020learning} & 2020 & TPAMI & node \& edge & & \checkmark & & \checkmark (JT) & feature addition\\\midrule

\multirow{15}{*}{\makecell{Graph Adversarial\\ Defense\\}} & Gasoline & \cite{xu2021graph} & 2022 & TheWebConf & node & \checkmark & \checkmark &  & \checkmark (BO) & \makecell{feature rewriting/graph rewiring}\\\cmidrule{2-11}
& Pro-GNN & \cite{jin2020graphkdd} & 2020 & KDD & node & \checkmark &  &  & \checkmark (JT) & graph rewiring\\\cmidrule{2-11}
& GIB-N & \cite{wu2020graphinfo} & 2020 & NeurIPS & node & \checkmark &  &  & \checkmark (JT) & graph rewiring\\\cmidrule{2-11}
& G-SVD & \cite{entezari2020all} & 2020 & WSDM & node & \checkmark & & & \checkmark (DT) & graph rewiring\\\cmidrule{2-11}
& RoGNN & \cite{wei2020decoupling} & 2020 & WCSP & node & \checkmark & & & \checkmark (JT) & graph rewiring \\\cmidrule{2-11}
& GNNGuard & \cite{zhang2020gnnguard} & 2020 & NeurIPS & node & \checkmark & & & \checkmark (JT) & graph rewiring \\\cmidrule{2-11}
& Flag & \cite{kong2020flag} & 2020 & arXiv & node \& edge \& graph & & \checkmark & & \checkmark (JT) & feature rewriting\\\cmidrule{2-11}
& G-Jaccard & \cite{wu2019adversarial} & 2019 & IJCAI & node & \checkmark & & & & graph rewiring\\\cmidrule{2-11}
& GraphAT & \cite{feng2019graph} & 2019 & TKDE & node & & \checkmark & & \checkmark (JT) & feature rewriting\\\midrule
\multirow{10}{*}{\makecell{Boosting GNN\\ Expressiveness}} & LAGNN & \cite{liu2022local} & 2022 & ICML & node \& edge \& graph & & \checkmark & & \checkmark (JT) & feature addition\\\cmidrule{2-11}
& rGINs & \cite{sato2021random} & 2021 & SDM & graph &  & \checkmark &  &  & feature addition\\\cmidrule{2-11}
& GSN & \cite{bouritsas2022improving} & 2021 & TPAMI & graph &  & \checkmark &  &  & feature addition\\\cmidrule{2-11}
& NGNN & \cite{zhang2021nested} & 2021 & NeurIPS & graph & \checkmark &  &  &  & graph sampling\\\cmidrule{2-11}
& ID-GNN & \cite{you2021identity} & 2021 & AAAI & node \& edge \& graph & \checkmark & \checkmark &  &  & \makecell{graph sampling/feature addition}\\\cmidrule{2-11}
& Distance Encoding & \cite{li2020distance} & 2020 & NeurIPS & node \& edge \& graph &  & \checkmark &  &  & feature addition \\\cmidrule{2-11}
& Master Node & \cite{gilmer2017neural} & 2017 & ICML & graph & \checkmark &  &  &  & node insertion\\\midrule

\multirow{8}{*}{\makecell{Alleviating\\ Over-Smoothing/\\Squashing}} & SDRF & \cite{topping2022understanding} & 2022 & ICLR & node & \checkmark &  &  &  & graph rewiring \\\cmidrule{2-11}
& ADC & \cite{zhao2021adaptive} & 2021 & NeurIPS & node & \checkmark &  &  & \checkmark (BO) & graph diffusion\\\cmidrule{2-11}
& SHADOW-GNN & \cite{zeng2021decoupling} & 2021 & NeurIPS & node \& edge & \checkmark &  &  &  & graph sampling\\\cmidrule{2-11}
& DropEdge & \cite{rong2019dropedge} & 2020 & ICLR & node & \checkmark &  &  &  & edge perturbation (dropping)\\\cmidrule{2-11}
& AdaEdge & \cite{chen2020measuring} & 2020 & AAAI & node & \checkmark &  &  & \checkmark (JT) & graph rewiring\\\cmidrule{2-11}
& GDC & \cite{klicpera2019diffusion} & 2019 & NeurIPS & node & \checkmark &  &  &  & graph diffusion\\

\midrule

\multirow{15}{*}{\makecell{Scalable GNN\\ Training}} 

& GCOND & \cite{jin2022graph} & 2022 & ICLR & node & \checkmark & \checkmark & \checkmark & \checkmark (BO) & graph generation\\\cmidrule{2-11}
& DosCond & \cite{jin2022condensing} & 2022 & KDD & node \& graph & \checkmark & \checkmark & \checkmark & \checkmark (BO) & graph generation\\\cmidrule{2-11}
& GOREN & \cite{cai2021graph} & 2021 & ICLR & node & \checkmark & \checkmark & \checkmark & \checkmark (DT) & graph generation\\\cmidrule{2-11}
& SpectralGC & \cite{jin2020graph} & 2020 & AISTATS & graph & \checkmark &  &  & \checkmark (DT) & graph generation\\\cmidrule{2-11}
& SIGN & \cite{rossi2020sign} & 2020 & arXiv & node & \checkmark &  &  &  & graph diffusion\\\cmidrule{2-11}
& PPRGO & \cite{bojchevski2020scaling} & 2020 & KDD & node & \checkmark &  &  &  & graph diffusion\\\cmidrule{2-11}
& GBP & \cite{chen2020scalable} & 2020 & NeurIPS & node & \checkmark &  &  &  & graph diffusion\\\cmidrule{2-11}
& GraphSAINT & \cite{zeng2019graphsaint} & 2020 & ICLR & node & \checkmark &  &  &  & graph sampling \\\cmidrule{2-11}
& LADIES & \cite{zou2019layer} & 2019 & NeurIPS & node & \checkmark &  &  &  & graph sampling\\\cmidrule{2-11}
& Cluster-GCN & \cite{chiang2019cluster} & 2019 & KDD & node & \checkmark &  &  & \checkmark (DT) & graph rewiring\\\cmidrule{2-11}
& Fast-GCN & \cite{chen2018fastgcn} & 2018 & ICLR & node & \checkmark &  &  &  & graph sampling\\\cmidrule{2-11}
& GraphSAGE & \cite{hamilton2017inductive} & 2017 & NIPS & node & \checkmark &  &  &  & graph sampling\\\bottomrule
\end{tabular}}
\end{table*}

\section{Graph Data Augmentation for \\Reliable Graph Learning}
\label{sec: optimal graph learning}

One main goal of GraphDA is to achieve \textit{Reliable Graph Learning} in the real-world scenarios by augmenting the input graph(s). Specifically, in this work we focus on improving the robustness, expressiveness, and scalability of DGL models via GraphDA for different challenging learning scenarios. We summarize the representative works in Table~\ref{tab:summary of robust graph learning}.








\smallskip
\noindent\textbf{Graph Structure Learning.} Due to various reasons such as fake connections~\cite{hooi2016fraudar}, over-personalized users~\cite{chen2020trading} and construction heuristics, the given graph structure is not optimal for downstream graph learning tasks. Graph structure learning is proposed as a solution for the above challenge. From a general sense, the core technique used for structure learning is \textbf{\textit{graph rewiring}}.

A series of methods rewire the given graphs following various node similarity metrics. In most cases, such metrics are learned from the given graph topology. For example, GAUG~\cite{zhao2021data} and IDGL~\cite{chen2020iterative} train edge predictors based on learned node embeddings. Besides, from the optimization perspective, it is feasible to directly incorporate the graph data (e.g., adjacency matrix) itself as a part of the optimization variables. Based on that, the \textbf{\textit{graph rewiring}} process is essentially guided by the optimization objective. TO-GNN~\cite{yang2019topology} is a representative work whose loss functions include smoothness-related regularizations and the update of graphs is gradient descent-based. In addition, instead of optimizing the graph itself, an interesting idea is to optimize the graph-related distributions (e.g., graph generation distributions and edge dropping distributions). After that, \textbf{\textit{graph rewiring}} can be conducted by sampling from those distributions. A representative work is LDS~\cite{franceschi2019learning} which assumes that every edge is sampled from an independent Bernoulli distribution. Other representative works in this line include Bayesian-GCNN~\cite{zhang2019bayesian}, GEN~\cite{wang2021graph}, NeuralSparse~\cite{zheng2020robust}, and PTDNet~\cite{luo2021learning}.

\smallskip
\noindent\textbf{Graph Feature Denoising.}
Compared to structure denoising, the research on graph feature denoising has received less attention. In general, most of the work is developed based on \textbf{\textit{feature rewriting}}. For instance, AirGNN~\cite{liu2021graph} regularizes the $l_{21}$ norm between the input node features and convoluted node features such that the model is more tolerant against abnormal features. To handle missing node features, a special case of suboptimal initial node features, \textbf{\textit{feature propagation}}~\cite{rossi2021unreasonable} diffuses the features from observed nodes to neighbors whose features are missing based on the heat diffusion equation; in other words, it imputes the missing node features with aggregated features from the neighborhood of the target nodes. GCNMF~\cite{taguchi2021graph} explicitly formulate the missing node features by Gaussian mixture models whose parameters are inferred from the downstream tasks. An effort named SAT~\cite{chen2020learning} reconstructs the missing features through the feature distribution, which is inferred from the topology distribution. To handle missing features on heterogeneous information networks, HGNN-AC~\cite{jin2021heterogeneous} imputes missing features from neighbor nodes' topology-based node embedding, while HGCA~\cite{he2022analyzing} designs a feature augmenter which is trained by maximizing the agreement between the augmented node embedding and the actual node embedding.

\smallskip
\noindent\textbf{Graph Adversarial Defense.}
Aside from the noise introduced in the data collection phase, GNNs are fragile against adversarial attacks on graph structure and features~\cite{sun2018adversarial}. Naturally, conducting GraphDA to recover and enhance (a part of) the poisoned graphs is effective to alleviate the performance degradation. In this section, we discuss recent advances in defending graph adversarial attacks via GraphDA.


Using prior knowledge about benign graphs (e.g., feature smoothness) to guide \textbf{\textit{graph rewiring}} is effective to defend graph adversarial attacks. 
For instance, works such as G-SVD~\cite{entezari2020all} and DefenseVGAE~\cite{zhang2020defensevgae} restore poisoned graphs into their reconstructed low-rank graphs which shows great empirical effectiveness again adversarial attacks. In addition,G-Jaccard~\cite{wu2019adversarial} and GNNGuard~\cite{zhang2020gnnguard} prune links whose head and tail nodes' feature similarity (or embedding similarity) is lower than a predefined threshold to eliminate potentially malicious edges. Alternatively, the graph prior knowledge can be realized by setting explicit regularization terms. Pro-GNN~\cite{jin2020graphkdd} includes the topology sparsity and feature smoothness regularization terms into the optimization objective which can guide the \textbf{\textit{graph rewiring}}. Besides, the supervision signals from downstream tasks can implicitly reflect the uncontaminated topology and feature distribution. For instance, Gasoline~\cite{xu2021graph} calibrates the given graph based on the evaluation performance (e.g., classification loss) of the validation nodes. GIB-N~\cite{wu2020graphinfo} adopts the information bottleneck objective~\cite{tishby2000information} which maximizes the mutual information between node labels and node embeddings and uses this objective function to guide the \textbf{\textit{graph rewiring}}.




It is worth noting that an established defense strategy named adversarial training~\cite{DBLP:journals/corr/GoodfellowSS14} is grafted onto the graph data. Examples include a family of graph (virtual) adversarial training~\cite{kong2020flag,feng2019graph,deng2019batch, wang2019graphdefense}. Their core idea is to adversarially conduct \textbf{\textit{edge perturbation}} and \textbf{\textit{feature corruption}} to generate various challenging graph data samples which maximize the classification loss. Then those samples are included in the training of downstream GNN models which can improve the robustness of GNNs against adversarial attacks.


\begin{figure}[t]
	\centering
	\scalebox{0.9}{
	\includegraphics[width=\columnwidth]{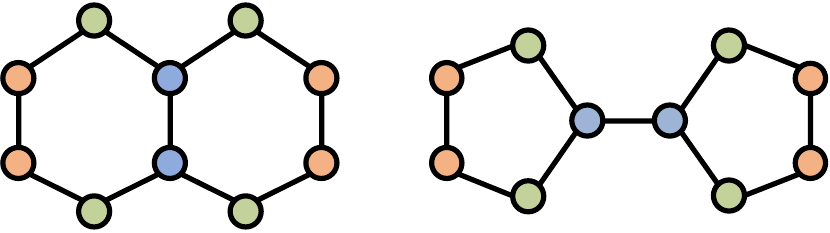}}
	\caption{1-WL test fails to distinguish the decalin graph (left) and the bicyclopentyl graph (right).}
	\vspace{-0.3cm}
	\label{fig:WL test}
\end{figure}

\smallskip
\noindent\textbf{Boosting GNN Expressive Power.}
Studies about the expressiveness of GNNs show that various mainstream GNNs cannot be more powerful than the 1-Weisfeiler-Lehman (WL) isomorphism test to distinguish graphs and subgraphs~\cite{morris2019weisfeiler,xu2018powerful}. E.g., Figure~\ref{fig:WL test} shows a pair of common examples that cannot be distinguished by the 1-WL test due to the common rooted trees. In addition, due to the power-law distribution of node degree, many nodes with few neighbors are hard to be represented properly. To break such an inherent limitation of GNNs, augmenting the given graphs is a feasible solution. One line of research adopts the \textbf{\textit{feature addition}} technique to augment the original node/edge features. For example, Distance Encoding~\cite{li2020distance} augments the given node features with distance measures between node pairs (or aggregated distance measures) to yield stronger expressiveness than 1-WL test-based GNNs. A recent study from Sato et al.~\cite{sato2021random} reveals that adding random features into the existing node features can boost the expressiveness of off-the-shelf GNNs (e.g., discriminate triangle structure). GSN~\cite{bouritsas2022improving} and a variant of ID-GNN~\cite{you2021identity} take a step further and augment the node features with the count of various motifs (e.g., cycles). Recently, LAGNN~\cite{liu2022local} adopts \textbf{\textit{feature addition}} by a trained feature generator from which nodes with few neighbors can benefit greatly.



Another line of work argues that the limitation of the 1-WL test (and 1-WL test-based GNNs) is related to the structure of rooted trees. Based on that, \textbf{\textit{graph sampling}} technique can enhance the GNN expressiveness. Specifically, NGNN~\cite{zhang2021nested} claims that node representations from sampled rooted subgraphs are more expressive than those from rooted trees. Interestingly, ID-GNN~\cite{you2021identity} shares a similar insight with NGNN~\cite{zhang2021nested} and also samples an ego network for every node for computing the node embedding.

As the node representations are obtained through the aggregation of node features within the receptive fields, to enhance the propagation of messages for long distance, a straightforward way is to apply \textbf{\textit{node insertion}} by inserting virtual nodes (i.e., super nodes or master nodes)~\cite{gilmer2017neural,li2017learning,ishiguro2019graph,pham2017graph} into the graphs which are connected with all the existing nodes.

\smallskip
\noindent\textbf{Alleviating Over-Smoothing/Squashing of GNNs.} Due to the inherent design limit of GNNs, as the depth of the model increases, the representations of different nodes in the graph eventually become indistinguishable with iterative message passing. The so-called over-smoothing phenomenon generally leads to failure in the graph learning tasks~\cite{li2018deeper}. To counter the over-smoothing issue, GraphDA, especially \textbf{\textit{graph rewiring}} methods have been shown as an effective solution. For example, DropEdge~\cite{rong2019dropedge} randomly removes graph edges during message passing to alleviate over-smoothing. TADropEdge~\cite{gao2021training} exploits graph structural information to compute edge weights for edge dropping, such that the augmented subgraphs can avoid the arbitrary data augmentation issue in DropEdge. AdaEdge~\cite{chen2020measuring} iteratively adds or removes edges to the graph topology based on the classification results (from a GNNs-based classifier) and trains GNN classifiers on the updated graphs to overcome the over-smoothing issue. SHADOW-GNN~\cite{zeng2021decoupling} samples informative subgraphs centered on each node and then builds a deep GNN operating on subgraphs instead of the whole graph to decouple the depth and scope of GNNs.

Note that over-smoothing is mostly demonstrated
in tasks that depend mostly on short-range information, while GNNs are also ineffective for capturing long-range node interactions. A recent study~\cite{alon2021bottleneck} points out that the distortion of information flowing from distant nodes (i.e., over-squashing) is the main factor limiting the efficiency of message passing for tasks relying on long-distance interactions. As the layer depth increases, the number of nodes in each node’s receptive field grows exponentially and leads to the over-squashing issue. To overcome over-squashing, \textbf{\textit{graph diffusion}} and \textbf{\textit{graph rewiring}} are two effective choices. Graph Diffusion Convolution (GDC)~\cite{klicpera2019diffusion} constructs a new graph based on a generalized form of graph diffusion, which can be further used to enlarge a larger neighborhood for message-passing. Adaptive Diffusion Convolution (ADC)~\cite{zhao2021adaptive} supports learning the optimal neighborhood from the data automatically to eliminate the manual search process of the optimal propagation neighborhood in GDC. Alon \& Yahav~\cite{alon2021bottleneck} propose to rewire the graph to build a fully-adjacent GNN layer for reducing the bottleneck. Similarly, Stochastic Discrete Ricci Flow (SDRF)~\cite{topping2022understanding} is a curvature-based method for graph rewiring to mitigate the over-squashing issue.

\smallskip
\noindent\textbf{Scalable Graph Training.}
Scalability is always a crucial challenge for GNNs which hinders the applicability of a broad class of GNN models over large real-world graphs. GraphDA techniques, such as \textbf{\textit{graph sampling}}, \textbf{\textit{graph diffusion}}, and \textbf{\textit{graph generation}} play an important role to speed up the training and inferring of GNNs.

Based on the idea of \textbf{\textit{graph sampling}}, GraphSAGE~\cite{hamilton2017inductive} uniformly samples the neighbors of the target nodes to enable GNN inductive learning on large graphs. FastGCN~\cite{chen2018fastgcn} shares a similar idea with GraphSAGE but follows the idea of importance sampling to sample vertices in every layer. To overcome the redundant sampling and sparse connection problem brought by the GraphSAGE~\cite{hamilton2017inductive} and FastGCN~\cite{chen2018fastgcn} respectively, Zou et al.~\cite{zou2019layer} propose LADIES which is a layer-dependent sampling strategy. In addition, GraphSAINT~\cite{zeng2019graphsaint} is another sampling-based method that applies various sampling methods (e.g., based on random walk) to improve the scalability of training rather than modifying the vanilla GCN model~\cite{kipf2016semi}. Similar to sampling methods, graph partition is also effective to lower the memory requirement. For example, Cluster-GCN~\cite{chiang2019cluster} partitions the given graph into clusters and conducts convolution operation within every cluster to avoid heavy neighborhood search and sampling. 

Another line of GraphDA research on scaling the training of GNNs adopts a decoupled design that trains MLP on the augmented graphs via \textbf{\textit{graph diffusion}} ~\cite{chen2020graph,wu2019simplifying,klicpera2018predict,rossi2020sign}. Given the above model design, the graph diffusion matrix can be pre-computed, which makes the training of MLP more efficient. A set of representative works adopt PageRank~\cite{page1999pagerank} and its variant (e.g., personalized PageRank~\cite{jeh2003scaling}) to infer the diffusion matrix such as SGC~\cite{wu2019simplifying} and APPNP~\cite{klicpera2018predict}. Similarly, PPRGo~\cite{bojchevski2020scaling} adopts the Forward Push algorithm~\cite{andersen2006local} to approximate the personalized PageRank matrix for efficiency. GBP~\cite{chen2020scalable} revisits the core idea of PPRGo and further speeds up the computation of the generalized PageRank matrix by the Bidirectional Propagation algorithm. In addition, a very recent work named GRAND+~\cite{feng2022grand+} also approximates the graph diffusion matrix to scale the training of graph random neural networks~\cite{feng2020graph} in the consistency training context.


In addition, works on ``graph condensation'' and ``graph coarsening'' try to shrink the given graph $\mathcal{G}$ by \textit{\textbf{graph generation}} so that the generated graph $\tilde{\mathcal{G}}$ can be handled by GNNs~\cite{huang2021scaling}. Their core idea is to minimize a `quantity of interest' between the input graph $\mathcal{G}$ and augmented graph $\tilde{\mathcal{G}}$. For example, Jin et al.~\cite{jin2020graph} propose to minimize the spectral distance between $\mathcal{G}$ and $\tilde{\mathcal{G}}$. GOREN~\cite{cai2021graph} aims to keep Laplace operators from $\mathcal{G}$ and $\tilde{\mathcal{G}}$ comparable. GCOND~\cite{jin2022graph} and DosCond~\cite{jin2022condensing} minimize the difference of training gradients on $\mathcal{G}$ and $\tilde{\mathcal{G}}$.




\section{Future Directions}
\label{sec:future challenges}
GraphDA is an emerging and fast-developing field. Although substantial progress has been achieved, many challenges remain under-explored. In this section, we discuss some promising research directions as follows.

\smallskip
\noindent\textbf{Data Augmentation beyond Simple Graphs.} Most of the aforementioned works develop augmentation strategies on homophilic (i.e., assortative) graphs where edges tend to connect nodes with the same properties (e.g., labels, features). However, heterophily (i.e., disassortativity) also exists commonly in networks such as heterosexual dating networks. Many existing augmentation approaches~\cite{DBLP:conf/kdd/SureshBNLM21} on heterophilic graphs focus on improving the assortativity of the given graphs or dropping/deweighting the existing disassortative edges~\cite{ye2021sparse}. The augmentation of node features, labels, and non-existing edges of heterophilic graphs remains understudied. Besides, existing GraphDA efforts are mainly developed for either plain or attributed graphs, while principled augmentation approaches for other types of graphs (e.g., heterogeneous graphs, hypergraphs, multiplex graphs, dynamic graphs) remain largely unexplored. Those complex graphs provide broader design space for augmentation but also challenge the effectiveness of existing GraphDA methods greatly, which is vital to explore in the future.



\smallskip
\noindent\textbf{Automated and Generalizable Graph Data Augmentation.}
In general, the effectiveness of DGL problems hinges on adhoc graph data augmentations, which have to be manually picked per dataset, by either rules of thumb or trial-and-errors. For example, researchers observe that different data augmentations affect downstream tasks differently across datasets, which suggests that searching over augmentation functions is crucial for graph self-supervised learning~\cite{DBLP:conf/icml/YouCSW21}. Nevertheless, evaluating representations derived from multiple augmentation functions without direct access to ground truth labels makes this problem challenging. Hence, it is necessary to develop automated data augmentation solutions to adaptively customize augmentation strategies for each graph dataset~\cite{luo2022automated}. Meanwhile, considering that different graphs usually have distinct graph properties, developing generalizable data augmentation methods without learning from scratch for each domain is also a promising direction to improve the practical usage of GraphDA methods.




\smallskip
\noindent\textbf{Semantic-Preserving Graph Data Augmentation.} Designing effective data augmentation for graphs is challenging due to their non-Euclidean nature and the dependencies between data samples. Most graph data augmentation methods adopt arbitrary augmentations on the input graph, which may unexpectedly change both structural and semantic patterns of the graph~\cite{park2021metropolis}. For example, dropping a carbon atom from the phenyl ring of aspirin breaks the aromatic system and results in a alkene chain, which is an entirely different chemical compound. Hence, proposing a label-consistent/semantic-preserving GraphDA method is of necessity. To this end, recent studies~\cite{lee2022augmentation,xia2022simgrace,yue2022label} perform data augmentation on the latent space to avoid the perturbation on the semantics.







\smallskip
\noindent\textbf{Graph Data Augmentation for Trustworthy DGL.} Despite the success of DGL, how to ensure various DGL algorithms behave in a socially responsible manner and meet regulatory compliance requirements becomes an emerging problem, especially in risk-sensitive applications. In fact, GraphDA can be an effective tool to achieve Trustworthy GML, especially on fairness, causality, explainability of DGL algorithms. For example, counterfactural graph data augmentation~\cite{lucic2022cf,zhao2022learning} has been used to explain the behavior of GNNs in the literature. Moreover, data augmentation itself is typically performed in an adhoc manner with little understanding of the underlying theoretical principles. Existing work on GraphDA is mainly surface-level, and rarely investigates the theoretical underpinnings and principles. Overall, there indeed appears to be a lack of research on interpret why exactly GraphDA works.


\section{Conclusion}


In this paper, we present a forward-looking and structured survey of graph data augmentation (GraphDA). To inspect the nature of GraphDA, we give a formal formulation and a taxonomy to facilitate the understanding of this emerging research problem. Specifically, we frame GraphDA methods into three categories according to the target augmentation modalities, i.e., feature-wise, structure-wise, and label-wise augmentations. We further review the application of GraphDA methods to address two data-centric DGL problems (i.e., low-resource graph learning and reliable graph learning) and discuss the prevailing GraphDA-based algorithms. Finally, we outline current challenges as well as opportunities for future research in this field.

\section*{Acknowledgments}
\thanks{This work is supported by NSF (No. 1947135,
2106825, 2229461, and 2134079), 
the NSF Program on Fairness in AI in collaboration with Amazon (No. 1939725), DARPA (No. HR001121C0165), ARO (No. W911NF2110088), ONR (No. N00014-21-1-4002), NIFA (No. 2020-67021-32799, and W911NF2110030), and ARL (No. W911NF2020124).}

%
\bibliographystyle{abbrv}
\bibliography{sigkddExp}  

\begin{thebibliography}{100}

\bibitem{alon2021bottleneck}
U.~Alon and E.~Yahav.
\newblock On the bottleneck of graph neural networks and its practical
  implications.
\newblock In {\em ICLR}, 2021.

\bibitem{andersen2006local}
R.~Andersen, F.~Chung, and K.~Lang.
\newblock Local graph partitioning using pagerank vectors.
\newblock In {\em FOCS}, 2006.

\bibitem{bishop1994neural}
C.~M. Bishop.
\newblock Neural networks and their applications.
\newblock {\em Review of scientific instruments}, 1994.

\bibitem{blum1998combining}
A.~Blum and T.~Mitchell.
\newblock Combining labeled and unlabeled data with co-training.
\newblock In {\em COLT}, 1998.

\bibitem{bo2022regularizing}
D.~Bo, B.~Hu, X.~Wang, Z.~Zhang, C.~Shi, and J.~Zhou.
\newblock Regularizing graph neural networks via consistency-diversity graph
  augmentations.
\newblock In {\em AAAI}, 2022.

\bibitem{bojchevski2020scaling}
A.~Bojchevski, J.~Klicpera, B.~Perozzi, A.~Kapoor, M.~Blais,
  B.~R{\'o}zemberczki, M.~Lukasik, and S.~G{\"u}nnemann.
\newblock Scaling graph neural networks with approximate pagerank.
\newblock In {\em KDD}, 2020.

\bibitem{bouritsas2022improving}
G.~Bouritsas, F.~Frasca, S.~P. Zafeiriou, and M.~Bronstein.
\newblock Improving graph neural network expressivity via subgraph isomorphism
  counting.
\newblock {\em TPAMI}, 2022.

\bibitem{cai2021graph}
C.~Cai, D.~Wang, and Y.~Wang.
\newblock Graph coarsening with neural networks.
\newblock In {\em ICLR}, 2021.

\bibitem{chen2020measuring}
D.~Chen, Y.~Lin, W.~Li, P.~Li, J.~Zhou, and X.~Sun.
\newblock Measuring and relieving the over-smoothing problem for graph neural
  networks from the topological view.
\newblock In {\em AAAI}, 2020.

\bibitem{chen2022graph}
H.~Chen, S.~Zhang, and G.~Xu.
\newblock Graph masked autoencoder.
\newblock {\em arXiv preprint arXiv:2202.08391}, 2022.

\bibitem{chen2018fastgcn}
J.~Chen, T.~Ma, and C.~Xiao.
\newblock Fastgcn: Fast learning with graph convolutional networks via
  importance sampling.
\newblock In {\em ICLR}, 2018.

\bibitem{chen2020graph}
L.~Chen, Z.~Chen, and J.~Bruna.
\newblock On graph neural networks versus graph-augmented mlps.
\newblock In {\em ICLR}, 2020.

\bibitem{chen2020trading}
L.~Chen, Y.~Yao, F.~Xu, M.~Xu, and H.~Tong.
\newblock Trading personalization for accuracy: Data debugging in collaborative
  filtering.
\newblock {\em NeurIPS}, 2020.

\bibitem{chen2020scalable}
M.~Chen, Z.~Wei, B.~Ding, Y.~Li, Y.~Yuan, X.~Du, and J.-R. Wen.
\newblock Scalable graph neural networks via bidirectional propagation.
\newblock {\em NeurIPS}, 2020.

\bibitem{chen2020learning}
X.~Chen, S.~Chen, J.~Yao, H.~Zheng, Y.~Zhang, and I.~W. Tsang.
\newblock Learning on attribute-missing graphs.
\newblock {\em TPAMI}, 2020.

\bibitem{chen2020iterative}
Y.~Chen, L.~Wu, and M.~Zaki.
\newblock Iterative deep graph learning for graph neural networks: Better and
  robust node embeddings.
\newblock {\em NeurIPS}, 2020.

\bibitem{chiang2019cluster}
W.-L. Chiang, X.~Liu, S.~Si, Y.~Li, S.~Bengio, and C.-J. Hsieh.
\newblock Cluster-gcn: An efficient algorithm for training deep and large graph
  convolutional networks.
\newblock In {\em KDD}, 2019.

\bibitem{dai2021nrgnn}
E.~Dai, C.~Aggarwal, and S.~Wang.
\newblock Nrgnn: Learning a label noise-resistant graph neural network on
  sparsely and noisily labeled graphs.
\newblock In {\em KDD}, 2021.

\bibitem{deng2019batch}
Z.~Deng, Y.~Dong, and J.~Zhu.
\newblock Batch virtual adversarial training for graph convolutional networks.
\newblock {\em arXiv preprint arXiv:1902.09192}, 2019.

\bibitem{ding2019deep}
K.~Ding, J.~Li, R.~Bhanushali, and H.~Liu.
\newblock Deep anomaly detection on attributed networks.
\newblock In {\em SDM}, 2019.

\bibitem{ding2022meta}
K.~Ding, J.~Wang, J.~Caverlee, and H.~Liu.
\newblock Meta propagation networks for graph few-shot semi-supervised
  learning.
\newblock In {\em AAAI}, 2022.

\bibitem{ding2020more}
K.~Ding, J.~Wang, J.~Li, D.~Li, and H.~Liu.
\newblock Be more with less: Hypergraph attention networks for inductive text
  classification.
\newblock In {\em EMNLP}, 2020.

\bibitem{ding2020graph}
K.~Ding, J.~Wang, J.~Li, K.~Shu, C.~Liu, and H.~Liu.
\newblock Graph prototypical networks for few-shot learning on attributed
  networks.
\newblock In {\em CIKM}, 2020.

\bibitem{ding2022structural}
K.~Ding, Y.~Wang, Y.~Yang, and H.~Liu.
\newblock Eliciting structural and semantic global knowledge in unsupervised
  graph contrastive learning.
\newblock In {\em AAAI}, 2023.

\bibitem{dong2021equivalence}
H.~Dong, J.~Chen, F.~Feng, X.~He, S.~Bi, Z.~Ding, and P.~Cui.
\newblock On the equivalence of decoupled graph convolution network and label
  propagation.
\newblock In {\em TheWebConf}, 2021.

\bibitem{entezari2020all}
N.~Entezari, S.~A. Al-Sayouri, A.~Darvishzadeh, and E.~E. Papalexakis.
\newblock All you need is low (rank) defending against adversarial attacks on
  graphs.
\newblock In {\em WSDM}, 2020.

\bibitem{feng2019graph}
F.~Feng, X.~He, J.~Tang, and T.-S. Chua.
\newblock Graph adversarial training: Dynamically regularizing based on graph
  structure.
\newblock {\em TKDE}, 2019.

\bibitem{feng2022adversarial}
S.~Feng, B.~Jing, Y.~Zhu, and H.~Tong.
\newblock Adversarial graph contrastive learning with information
  regularization.
\newblock In {\em TheWebConf}, 2022.

\bibitem{feng2022grand+}
W.~Feng, Y.~Dong, T.~Huang, Z.~Yin, X.~Cheng, E.~Kharlamov, and J.~Tang.
\newblock Grand+: Scalable graph random neural networks.
\newblock In {\em TheWebConf}, 2022.

\bibitem{feng2020graph}
W.~Feng, J.~Zhang, Y.~Dong, Y.~Han, H.~Luan, Q.~Xu, Q.~Yang, E.~Kharlamov, and
  J.~Tang.
\newblock Graph random neural networks for semi-supervised learning on graphs.
\newblock In {\em NeurIPS}, 2020.

\bibitem{franceschi2019learning}
L.~Franceschi, M.~Niepert, M.~Pontil, and X.~He.
\newblock Learning discrete structures for graph neural networks.
\newblock In {\em ICML}, 2019.

\bibitem{gao2021training}
Z.~Gao, S.~Bhattacharya, L.~Zhang, R.~S. Blum, A.~Ribeiro, and B.~M. Sadler.
\newblock Training robust graph neural networks with topology adaptive edge
  dropping.
\newblock {\em arXiv preprint arXiv:2106.02892}, 2021.

\bibitem{gilmer2017neural}
J.~Gilmer, S.~S. Schoenholz, P.~F. Riley, O.~Vinyals, and G.~E. Dahl.
\newblock Neural message passing for quantum chemistry.
\newblock In {\em ICML}, 2017.

\bibitem{DBLP:journals/corr/GoodfellowSS14}
I.~J. Goodfellow, J.~Shlens, and C.~Szegedy.
\newblock Explaining and harnessing adversarial examples.
\newblock In {\em ICLR}, 2015.

\bibitem{guo2021intrusion}
H.~Guo and Y.~Mao.
\newblock ifmixup: Towards intrusion-free graph mixup for graph classification.
\newblock {\em arXiv preprint arXiv:2110.09344}, 2021.

\bibitem{hamilton2017inductive}
W.~L. Hamilton, R.~Ying, and J.~Leskovec.
\newblock Inductive representation learning on large graphs.
\newblock {\em arXiv preprint arXiv:1706.02216}, 2017.

\bibitem{han2022g}
X.~Han, Z.~Jiang, N.~Liu, and X.~Hu.
\newblock G-mixup: Graph data augmentation for graph classification.
\newblock In {\em ICML}, 2022.

\bibitem{hassani2020contrastive}
K.~Hassani and A.~H. Khasahmadi.
\newblock Contrastive multi-view representation learning on graphs.
\newblock In {\em ICML}, 2020.

\bibitem{hassani2022learning}
K.~Hassani and A.~H. Khasahmadi.
\newblock Learning graph augmentations to learn graph representations.
\newblock {\em arXiv preprint arXiv:2201.09830}, 2022.

\bibitem{hawkins2021scalable}
C.~Hawkins, V.~N. Ioannidis, S.~Adeshina, and G.~Karypis.
\newblock Scalable consistency training for graph neural networks via
  self-ensemble self-distillation.
\newblock {\em arXiv preprint arXiv:2110.06290}, 2021.

\bibitem{he2022analyzing}
D.~He, C.~Liang, C.~Huo, Z.~Feng, D.~Jin, L.~Yang, and W.~Zhang.
\newblock Analyzing heterogeneous networks with missing attributes by
  unsupervised contrastive learning.
\newblock {\em TNNLS}, 2022.

\bibitem{hooi2016fraudar}
B.~Hooi, H.~A. Song, A.~Beutel, N.~Shah, K.~Shin, and C.~Faloutsos.
\newblock Fraudar: Bounding graph fraud in the face of camouflage.
\newblock In {\em KDD}, 2016.

\bibitem{hou2022graphmae}
Z.~Hou, X.~Liu, Y.~Dong, C.~Wang, J.~Tang, et~al.
\newblock Graphmae: Self-supervised masked graph autoencoders.
\newblock In {\em KDD}, 2022.

\bibitem{hu2020gpt}
Z.~Hu, Y.~Dong, K.~Wang, K.-W. Chang, and Y.~Sun.
\newblock Gpt-gnn: Generative pre-training of graph neural networks.
\newblock In {\em KDD}, 2020.

\bibitem{hu2019pre}
Z.~Hu, C.~Fan, T.~Chen, K.-W. Chang, and Y.~Sun.
\newblock Pre-training graph neural networks for generic structural feature
  extraction.
\newblock {\em arXiv preprint arXiv:1905.13728}, 2019.

\bibitem{huang2021scaling}
Z.~Huang, S.~Zhang, C.~Xi, T.~Liu, and M.~Zhou.
\newblock Scaling up graph neural networks via graph coarsening.
\newblock In {\em KDD}, 2021.

\bibitem{hui2020collaborative}
B.~Hui, P.~Zhu, and Q.~Hu.
\newblock Collaborative graph convolutional networks: Unsupervised learning
  meets semi-supervised learning.
\newblock In {\em AAAI}, 2020.

\bibitem{ishiguro2019graph}
K.~Ishiguro, S.-i. Maeda, and M.~Koyama.
\newblock Graph warp module: an auxiliary module for boosting the power of
  graph neural networks in molecular graph analysis.
\newblock {\em arXiv preprint arXiv:1902.01020}, 2019.

\bibitem{jeh2003scaling}
G.~Jeh and J.~Widom.
\newblock Scaling personalized web search.
\newblock In {\em WWW}, 2003.

\bibitem{jiao2020sub}
Y.~Jiao, Y.~Xiong, J.~Zhang, Y.~Zhang, T.~Zhang, and Y.~Zhu.
\newblock Sub-graph contrast for scalable self-supervised graph representation
  learning.
\newblock In {\em ICDM}, 2020.

\bibitem{jin2021heterogeneous}
D.~Jin, C.~Huo, C.~Liang, and L.~Yang.
\newblock Heterogeneous graph neural network via attribute completion.
\newblock In {\em TheWebConf}, 2021.

\bibitem{jin2021multi}
M.~Jin, Y.~Zheng, Y.-F. Li, C.~Gong, C.~Zhou, and S.~Pan.
\newblock Multi-scale contrastive siamese networks for self-supervised graph
  representation learning.
\newblock In {\em IJCAI}, 2021.

\bibitem{jin2020graphkdd}
W.~Jin, Y.~Ma, X.~Liu, X.~Tang, S.~Wang, and J.~Tang.
\newblock Graph structure learning for robust graph neural networks.
\newblock In {\em KDD}, 2020.

\bibitem{jin2022condensing}
W.~Jin, X.~Tang, H.~Jiang, Z.~Li, D.~Zhang, J.~Tang, and B.~Yin.
\newblock Condensing graphs via one-step gradient matching.
\newblock In {\em KDD}, 2022.

\bibitem{jin2022graph}
W.~Jin, L.~Zhao, S.~Zhang, Y.~Liu, J.~Tang, and N.~Shah.
\newblock Graph condensation for graph neural networks.
\newblock In {\em ICLR}, 2022.

\bibitem{jin2020graph}
Y.~Jin, A.~Loukas, and J.~JaJa.
\newblock Graph coarsening with preserved spectral properties.
\newblock In {\em AISTATS}, 2020.

\bibitem{kazi2022differentiable}
A.~Kazi, L.~Cosmo, S.-A. Ahmadi, N.~Navab, and M.~Bronstein.
\newblock Differentiable graph module (dgm) for graph convolutional networks.
\newblock {\em TPAMI}, 2022.

\bibitem{kipf2016variational}
T.~N. Kipf and M.~Welling.
\newblock Variational graph auto-encoders.
\newblock {\em arXiv preprint arXiv:1611.07308}, 2016.

\bibitem{kipf2016semi}
T.~N. Kipf and M.~Welling.
\newblock Semi-supervised classification with graph convolutional networks.
\newblock In {\em ICLR}, 2017.

\bibitem{klicpera2018predict}
J.~Klicpera, A.~Bojchevski, and S.~G{\"u}nnemann.
\newblock Predict then propagate: Graph neural networks meet personalized
  pagerank.
\newblock In {\em ICLR}, 2018.

\bibitem{klicpera2019diffusion}
J.~Klicpera, S.~Wei{\ss}enberger, and S.~G{\"u}nnemann.
\newblock Diffusion improves graph learning.
\newblock In {\em NeurIPS}, 2019.

\bibitem{kondor2002diffusion}
R.~I. Kondor and J.~Lafferty.
\newblock Diffusion kernels on graphs and other discrete structures.
\newblock In {\em ICML}, 2002.

\bibitem{kong2020flag}
K.~Kong, G.~Li, M.~Ding, Z.~Wu, C.~Zhu, B.~Ghanem, G.~Taylor, and T.~Goldstein.
\newblock Flag: Adversarial data augmentation for graph neural networks.
\newblock {\em arXiv preprint arXiv:2010.09891}, 2020.

\bibitem{lee2022augmentation}
N.~Lee, J.~Lee, and C.~Park.
\newblock Augmentation-free self-supervised learning on graphs.
\newblock In {\em AAAI}, 2022.

\bibitem{li2017learning}
J.~Li, D.~Cai, and X.~He.
\newblock Learning graph-level representation for drug discovery.
\newblock {\em arXiv preprint arXiv:1709.03741}, 2017.

\bibitem{li2020distance}
P.~Li, Y.~Wang, H.~Wang, and J.~Leskovec.
\newblock Distance encoding: Design provably more powerful neural networks for
  graph representation learning.
\newblock {\em NeurIPS}, 2020.

\bibitem{li2018deeper}
Q.~Li, Z.~Han, and X.-M. Wu.
\newblock Deeper insights into graph convolutional networks for semi-supervised
  learning.
\newblock In {\em AAAI}, 2018.

\bibitem{li2020co}
S.~Li, W.-T. Li, and W.~Wang.
\newblock Co-gcn for multi-view semi-supervised learning.
\newblock In {\em AAAI}, 2020.

\bibitem{li2022graph}
X.~Li, L.~Wen, Y.~Deng, F.~Feng, X.~Hu, L.~Wang, and Z.~Fan.
\newblock Graph neural network with curriculum learning for imbalanced node
  classification.
\newblock {\em arXiv preprint arXiv:2202.02529}, 2022.

\bibitem{liu2022local}
S.~Liu, H.~Dong, L.~Li, T.~Xu, Y.~Rong, P.~Zhao, J.~Huang, and D.~Wu.
\newblock Local augmentation for graph neural networks.
\newblock {\em ICML}, 2022.

\bibitem{liu2021graph}
X.~Liu, J.~Ding, W.~Jin, H.~Xu, Y.~Ma, Z.~Liu, and J.~Tang.
\newblock Graph neural networks with adaptive residual.
\newblock {\em NeurIPS}, 2021.

\bibitem{liu2022graph}
Y.~Liu, M.~Jin, S.~Pan, C.~Zhou, Y.~Zheng, F.~Xia, and P.~Yu.
\newblock Graph self-supervised learning: A survey.
\newblock {\em TKDE}, 2022.

\bibitem{liu2022gatsmote}
Y.~Liu, Z.~Zhang, Y.~Liu, and Y.~Zhu.
\newblock Gatsmote: Improving imbalanced node classification on graphs via
  attention and homophily.
\newblock {\em Mathematics}, 2022.

\bibitem{lucic2022cf}
A.~Lucic, M.~A. Ter~Hoeve, G.~Tolomei, M.~De~Rijke, and F.~Silvestri.
\newblock Cf-gnnexplainer: Counterfactual explanations for graph neural
  networks.
\newblock In {\em AISTATS}, 2022.

\bibitem{luo2021learning}
D.~Luo, W.~Cheng, W.~Yu, B.~Zong, J.~Ni, H.~Chen, and X.~Zhang.
\newblock Learning to drop: Robust graph neural network via topological
  denoising.
\newblock In {\em WSDM}, 2021.

\bibitem{luo2022automated}
Y.~Luo, M.~McThrow, W.~Y. Au, T.~Komikado, K.~Uchino, K.~Maruhash, and S.~Ji.
\newblock Automated data augmentations for graph classification.
\newblock {\em arXiv preprint arXiv:2202.13248}, 2022.

\bibitem{mahmood2021masked}
O.~Mahmood, E.~Mansimov, R.~Bonneau, and K.~Cho.
\newblock Masked graph modeling for molecule generation.
\newblock {\em Nature communications}, 2021.

\bibitem{mo2022simple}
Y.~Mo, L.~Peng, J.~Xu, X.~Shi, and X.~Zhu.
\newblock Simple unsupervised graph representation learning.
\newblock In {\em AAAI}, 2022.

\bibitem{morris2019weisfeiler}
C.~Morris, M.~Ritzert, M.~Fey, W.~L. Hamilton, J.~E. Lenssen, G.~Rattan, and
  M.~Grohe.
\newblock Weisfeiler and leman go neural: Higher-order graph neural networks.
\newblock In {\em AAAI}, 2019.

\bibitem{page1999pagerank}
L.~Page, S.~Brin, R.~Motwani, and T.~Winograd.
\newblock The pagerank citation ranking: Bringing order to the web.
\newblock Technical report, Stanford InfoLab, 1999.

\bibitem{park2021metropolis}
H.~Park, S.~Lee, S.~Kim, J.~Park, J.~Jeong, K.-M. Kim, J.-W. Ha, and H.~J. Kim.
\newblock Metropolis-hastings data augmentation for graph neural networks.
\newblock {\em NeurIPS}, 2021.

\bibitem{park2022graph}
J.~Park, H.~Shim, and E.~Yang.
\newblock Graph transplant: Node saliency-guided graph mixup with local
  structure preservation.
\newblock In {\em AAAI}, 2022.

\bibitem{pham2017graph}
T.~Pham, T.~Tran, H.~Dam, and S.~Venkatesh.
\newblock Graph classification via deep learning with virtual nodes.
\newblock {\em arXiv preprint arXiv:1708.04357}, 2017.

\bibitem{qiu2020gcc}
J.~Qiu, Q.~Chen, Y.~Dong, J.~Zhang, H.~Yang, M.~Ding, K.~Wang, and J.~Tang.
\newblock Gcc: Graph contrastive coding for graph neural network pre-training.
\newblock In {\em KDD}, 2020.

\bibitem{rong2019dropedge}
Y.~Rong, W.~Huang, T.~Xu, and J.~Huang.
\newblock Dropedge: Towards deep graph convolutional networks on node
  classification.
\newblock In {\em ICLR}, 2019.

\bibitem{rossi2020sign}
E.~Rossi, F.~Frasca, B.~Chamberlain, D.~Eynard, M.~Bronstein, and F.~Monti.
\newblock Sign: Scalable inception graph neural networks.
\newblock {\em arXiv preprint arXiv:2004.11198}, 2020.

\bibitem{rossi2021unreasonable}
E.~Rossi, H.~Kenlay, M.~I. Gorinova, B.~P. Chamberlain, X.~Dong, and
  M.~Bronstein.
\newblock On the unreasonable effectiveness of feature propagation in learning
  on graphs with missing node features.
\newblock {\em arXiv preprint arXiv:2111.12128}, 2021.

\bibitem{sato2021random}
R.~Sato, M.~Yamada, and H.~Kashima.
\newblock Random features strengthen graph neural networks.
\newblock In {\em SDM}, 2021.

\bibitem{shorten2019survey}
C.~Shorten and T.~M. Khoshgoftaar.
\newblock A survey on image data augmentation for deep learning.
\newblock {\em Journal of Big Data}, 2019.

\bibitem{sun2020multi}
K.~Sun, Z.~Lin, and Z.~Zhu.
\newblock Multi-stage self-supervised learning for graph convolutional networks
  on graphs with few labeled nodes.
\newblock In {\em AAAI}, 2020.

\bibitem{sun2018adversarial}
L.~Sun, Y.~Dou, C.~Yang, J.~Wang, P.~S. Yu, L.~He, and B.~Li.
\newblock Adversarial attack and defense on graph data: A survey.
\newblock {\em arXiv preprint arXiv:1812.10528}, 2018.

\bibitem{DBLP:conf/kdd/SureshBNLM21}
S.~Suresh, V.~Budde, J.~Neville, P.~Li, and J.~Ma.
\newblock Breaking the limit of graph neural networks by improving the
  assortativity of graphs with local mixing patterns.
\newblock In {\em KDD}, 2021.

\bibitem{suresh2021adversarial}
S.~Suresh, P.~Li, C.~Hao, and J.~Neville.
\newblock Adversarial graph augmentation to improve graph contrastive learning.
\newblock In {\em NeurIPS}, 2021.

\bibitem{taguchi2021graph}
H.~Taguchi, X.~Liu, and T.~Murata.
\newblock Graph convolutional networks for graphs containing missing features.
\newblock {\em FGCS}, 2021.

\bibitem{tan2022mgae}
Q.~Tan, N.~Liu, X.~Huang, R.~Chen, S.-H. Choi, and X.~Hu.
\newblock Mgae: Masked autoencoders for self-supervised learning on graphs.
\newblock {\em arXiv preprint arXiv:2201.02534}, 2022.

\bibitem{thakoor2021large}
S.~Thakoor, C.~Tallec, M.~G. Azar, M.~Azabou, E.~L. Dyer, R.~Munos,
  P.~Veli{\v{c}}kovi{\'c}, and M.~Valko.
\newblock Large-scale representation learning on graphs via bootstrapping.
\newblock In {\em ICLR}, 2021.

\bibitem{tishby2000information}
N.~Tishby, F.~C. Pereira, and W.~Bialek.
\newblock The information bottleneck method.
\newblock {\em arXiv preprint physics/0004057}, 2000.

\bibitem{topping2022understanding}
J.~Topping, F.~Di~Giovanni, B.~P. Chamberlain, X.~Dong, and M.~M. Bronstein.
\newblock Understanding over-squashing and bottlenecks on graphs via curvature.
\newblock In {\em ICLR}, 2022.

\bibitem{velickovic2019deep}
P.~Velickovic, W.~Fedus, W.~L. Hamilton, P.~Li{\`o}, Y.~Bengio, and R.~D.
  Hjelm.
\newblock Deep graph infomax.
\newblock {\em ICLR}, 2019.

\bibitem{verma2019manifold}
V.~Verma, A.~Lamb, C.~Beckham, A.~Najafi, I.~Mitliagkas, D.~Lopez-Paz, and
  Y.~Bengio.
\newblock Manifold mixup: Better representations by interpolating hidden
  states.
\newblock In {\em ICML}, 2019.

\bibitem{verma2021graphmix}
V.~Verma, M.~Qu, K.~Kawaguchi, A.~Lamb, Y.~Bengio, J.~Kannala, and J.~Tang.
\newblock Graphmix: Improved training of gnns for semi-supervised learning.
\newblock In {\em AAAI}, 2021.

\bibitem{wang2020next}
J.~Wang, K.~Ding, L.~Hong, H.~Liu, and J.~Caverlee.
\newblock Next-item recommendation with sequential hypergraphs.
\newblock In {\em SIGIR}, 2020.

\bibitem{wang2021graph}
R.~Wang, S.~Mou, X.~Wang, W.~Xiao, Q.~Ju, C.~Shi, and X.~Xie.
\newblock Graph structure estimation neural networks.
\newblock In {\em TheWebConf}, 2021.

\bibitem{wang2021dual}
X.~Wang and J.~Chen.
\newblock A dual-branch graph convolutional network on imbalanced node
  classification.
\newblock In {\em CSAE}, 2021.

\bibitem{wang2019graphdefense}
X.~Wang, X.~Liu, and C.-J. Hsieh.
\newblock Graphdefense: Towards robust graph convolutional networks.
\newblock {\em arXiv preprint arXiv:1911.04429}, 2019.

\bibitem{wang2021distance}
Y.~Wang, C.~Aggarwal, and T.~Derr.
\newblock Distance-wise prototypical graph neural network in node imbalance
  classification.
\newblock {\em arXiv preprint arXiv:2110.12035}, 2021.

\bibitem{wang2021mixup}
Y.~Wang, W.~Wang, Y.~Liang, Y.~Cai, and B.~Hooi.
\newblock Mixup for node and graph classification.
\newblock In {\em TheWebConf}, 2021.

\bibitem{wang2020nodeaug}
Y.~Wang, W.~Wang, Y.~Liang, Y.~Cai, J.~Liu, and B.~Hooi.
\newblock Nodeaug: Semi-supervised node classification with data augmentation.
\newblock In {\em KDD}, 2020.

\bibitem{wei2020decoupling}
X.~Wei, Y.~Li, X.~Qin, X.~Xu, X.~Li, and M.~Liu.
\newblock From decoupling to reconstruction: A robust graph neural network
  against topology attacks.
\newblock In {\em WCSP}, 2020.

\bibitem{wu2019simplifying}
F.~Wu, A.~Souza, T.~Zhang, C.~Fifty, T.~Yu, and K.~Weinberger.
\newblock Simplifying graph convolutional networks.
\newblock In {\em ICML}, 2019.

\bibitem{wu2019adversarial}
H.~Wu, C.~Wang, Y.~Tyshetskiy, A.~Docherty, K.~Lu, and L.~Zhu.
\newblock Adversarial examples for graph data: deep insights into attack and
  defense.
\newblock In {\em IJCAI}, 2019.

\bibitem{wu2021graphmixup}
L.~Wu, H.~Lin, Z.~Gao, C.~Tan, S.~Li, et~al.
\newblock Graphmixup: Improving class-imbalanced node classification on graphs
  by self-supervised context prediction.
\newblock {\em arXiv preprint arXiv:2106.11133}, 2021.

\bibitem{wu2021self}
L.~Wu, H.~Lin, C.~Tan, Z.~Gao, and S.~Z. Li.
\newblock Self-supervised learning on graphs: Contrastive, generative, or
  predictive.
\newblock {\em TKDE}, 2021.

\bibitem{wu2020graphinfo}
T.~Wu, H.~Ren, P.~Li, and J.~Leskovec.
\newblock Graph information bottleneck.
\newblock In {\em NeurIPS}, 2020.

\bibitem{wu2018quest}
X.~Wu, L.~Zhao, and L.~Akoglu.
\newblock A quest for structure: jointly learning the graph structure and
  semi-supervised classification.
\newblock In {\em CIKM}, 2018.

\bibitem{wu2020comprehensive}
Z.~Wu, S.~Pan, F.~Chen, G.~Long, C.~Zhang, and S.~Y. Philip.
\newblock A comprehensive survey on graph neural networks.
\newblock {\em TNNLS}, 2020.

\bibitem{xia2022simgrace}
J.~Xia, L.~Wu, J.~Chen, B.~Hu, and S.~Z. Li.
\newblock Simgrace: A simple framework for graph contrastive learning without
  data augmentation.
\newblock In {\em TheWebConf}, 2022.

\bibitem{xie2020unsupervised}
Q.~Xie, Z.~Dai, E.~Hovy, T.~Luong, and Q.~Le.
\newblock Unsupervised data augmentation for consistency training.
\newblock In {\em NeurIPS}, 2020.

\bibitem{xie2022self}
Y.~Xie, Z.~Xu, J.~Zhang, Z.~Wang, and S.~Ji.
\newblock Self-supervised learning of graph neural networks: A unified review.
\newblock {\em TPAMI}, 2022.

\bibitem{xu2018powerful}
K.~Xu, W.~Hu, J.~Leskovec, and S.~Jegelka.
\newblock How powerful are graph neural networks?
\newblock In {\em ICLR}, 2018.

\bibitem{xu2021graph}
Z.~Xu, B.~Du, and H.~Tong.
\newblock Graph sanitation with application to node classification.
\newblock {\em arXiv preprint arXiv:2105.09384}, 2021.

\bibitem{yang2019topology}
L.~Yang, Z.~Kang, X.~Cao, D.~Jin, B.~Yang, and Y.~Guo.
\newblock Topology optimization based graph convolutional network.
\newblock In {\em IJCAI}, 2019.

\bibitem{yang2021graph}
L.~Yang, L.~Zhang, and W.~Yang.
\newblock Graph adversarial self-supervised learning.
\newblock In {\em NeurIPS}, 2021.

\bibitem{yarowsky1995unsupervised}
D.~Yarowsky.
\newblock Unsupervised word sense disambiguation rivaling supervised methods.
\newblock In {\em ACL}, 1995.

\bibitem{ye2021sparse}
Y.~Ye and S.~Ji.
\newblock Sparse graph attention networks.
\newblock {\em TKDE}, 2021.

\bibitem{you2021identity}
J.~You, J.~M. Gomes-Selman, R.~Ying, and J.~Leskovec.
\newblock Identity-aware graph neural networks.
\newblock In {\em AAAI}, 2021.

\bibitem{DBLP:conf/icml/YouCSW21}
Y.~You, T.~Chen, Y.~Shen, and Z.~Wang.
\newblock Graph contrastive learning automated.
\newblock In {\em ICML}, 2021.

\bibitem{you2020graph}
Y.~You, T.~Chen, Y.~Sui, T.~Chen, Z.~Wang, and Y.~Shen.
\newblock Graph contrastive learning with augmentations.
\newblock In {\em NeurIPS}, 2020.

\bibitem{you2020does}
Y.~You, T.~Chen, Z.~Wang, and Y.~Shen.
\newblock When does self-supervision help graph convolutional networks?
\newblock In {\em ICML}, 2020.

\bibitem{yue2022label}
H.~Yue, C.~Zhang, C.~Zhang, and H.~Liu.
\newblock Label-invariant augmentation for semi-supervised graph
  classification.
\newblock In {\em NeurIPS}, 2022.

\bibitem{zeng2021decoupling}
H.~Zeng, M.~Zhang, Y.~Xia, A.~Srivastava, A.~Malevich, R.~Kannan, V.~Prasanna,
  L.~Jin, and R.~Chen.
\newblock Decoupling the depth and scope of graph neural networks.
\newblock {\em NeurIPS}, 2021.

\bibitem{zeng2019graphsaint}
H.~Zeng, H.~Zhou, A.~Srivastava, R.~Kannan, and V.~Prasanna.
\newblock Graphsaint: Graph sampling based inductive learning method.
\newblock In {\em ICLR}, 2019.

\bibitem{zeng2021contrastive}
J.~Zeng and P.~Xie.
\newblock Contrastive self-supervised learning for graph classification.
\newblock In {\em AAAI}, 2021.

\bibitem{zhang2020defensevgae}
A.~Zhang and J.~Ma.
\newblock Defensevgae: Defending against adversarial attacks on graph data via
  a variational graph autoencoder.
\newblock {\em arXiv preprint arXiv:2006.08900}, 2020.

\bibitem{zhang2021improving}
C.~Zhang, Y.~He, Y.~Cen, Z.~Hou, and J.~Tang.
\newblock Improving the training of graph neural networks with consistency
  regularization.
\newblock {\em arXiv preprint arXiv:2112.04319}, 2021.

\bibitem{zhang2018mixup}
H.~Zhang, M.~Cisse, Y.~N. Dauphin, and D.~Lopez-Paz.
\newblock mixup: Beyond empirical risk minimization.
\newblock In {\em ICLR}, 2018.

\bibitem{zhang2020graph}
J.~Zhang, H.~Zhang, C.~Xia, and L.~Sun.
\newblock Graph-bert: Only attention is needed for learning graph
  representations.
\newblock {\em arXiv preprint arXiv:2001.05140}, 2020.

\bibitem{zhang2021nested}
M.~Zhang and P.~Li.
\newblock Nested graph neural networks.
\newblock {\em NeurIPS}, 2021.

\bibitem{zhang2019graph}
S.~Zhang, H.~Tong, J.~Xu, and R.~Maciejewski.
\newblock Graph convolutional networks: a comprehensive review.
\newblock {\em Computational Social Networks}, 2019.

\bibitem{zhang2020gnnguard}
X.~Zhang and M.~Zitnik.
\newblock Gnnguard: Defending graph neural networks against adversarial
  attacks.
\newblock {\em NeurIPS}, 2020.

\bibitem{zhang2019bayesian}
Y.~Zhang, S.~Pal, M.~Coates, and D.~Ustebay.
\newblock Bayesian graph convolutional neural networks for semi-supervised
  classification.
\newblock In {\em AAAI}, 2019.

\bibitem{zhao2021adaptive}
J.~Zhao, Y.~Dong, M.~Ding, E.~Kharlamov, and J.~Tang.
\newblock Adaptive diffusion in graph neural networks.
\newblock In {\em NeurIPS}, 2021.

\bibitem{zhao2021heterogeneous}
J.~Zhao, X.~Wang, C.~Shi, B.~Hu, G.~Song, and Y.~Ye.
\newblock Heterogeneous graph structure learning for graph neural networks.
\newblock In {\em AAAI}, 2021.

\bibitem{zhao2022learning}
T.~Zhao, G.~Liu, D.~Wang, W.~Yu, and M.~Jiang.
\newblock Learning from counterfactual links for link prediction.
\newblock In {\em ICML}, 2022.

\bibitem{zhao2021data}
T.~Zhao, Y.~Liu, L.~Neves, O.~Woodford, M.~Jiang, and N.~Shah.
\newblock Data augmentation for graph neural networks.
\newblock In {\em AAAI}, 2021.

\bibitem{zhao2021graphsmote}
T.~Zhao, X.~Zhang, and S.~Wang.
\newblock Graphsmote: Imbalanced node classification on graphs with graph
  neural networks.
\newblock In {\em WSDM}, 2021.

\bibitem{zheng2020robust}
C.~Zheng, B.~Zong, W.~Cheng, D.~Song, J.~Ni, W.~Yu, H.~Chen, and W.~Wang.
\newblock Robust graph representation learning via neural sparsification.
\newblock In {\em ICML}, 2020.

\bibitem{zhu2020deep}
Y.~Zhu, Y.~Xu, F.~Yu, Q.~Liu, S.~Wu, and L.~Wang.
\newblock Deep graph contrastive representation learning.
\newblock {\em arXiv preprint arXiv:2006.04131}, 2020.

\bibitem{zhu2021graph}
Y.~Zhu, Y.~Xu, F.~Yu, Q.~Liu, S.~Wu, and L.~Wang.
\newblock Graph contrastive learning with adaptive augmentation.
\newblock In {\em TheWebConf}, 2021.

\bibitem{zou2019layer}
D.~Zou, Z.~Hu, Y.~Wang, S.~Jiang, Y.~Sun, and Q.~Gu.
\newblock Layer-dependent importance sampling for training deep and large graph
  convolutional networks.
\newblock {\em NeurIPS}, 2019.

\end{thebibliography}

%
%

\end{document}